\documentclass{article}

\usepackage{microtype}
\usepackage{graphicx}
\usepackage{subcaption}
\usepackage{booktabs} % for professional tables
\usepackage{svg}

\usepackage{hyperref}

\usepackage[accepted]{icml2026}

\usepackage{amsmath}
\usepackage{amssymb}
\usepackage{mathtools}
\usepackage{amsthm}
\usepackage{pifont}
\usepackage{enumitem}

\usepackage[capitalize,noabbrev]{cleveref}
\usepackage{xspace}
\usepackage{amsmath,amsfonts,bm}

\def\eqref#1{equation~\ref{#1}}
\def\1{\bm{1}}

\def\vk{{\bm{k}}}

\def\vq{{\bm{q}}}

\def\vv{{\bm{v}}}

\def\mB{{\bm{B}}}

\def\mM{{\bm{M}}}

\DeclareMathAlphabet{\mathsfit}{\encodingdefault}{\sfdefault}{m}{sl}
\SetMathAlphabet{\mathsfit}{bold}{\encodingdefault}{\sfdefault}{bx}{n}

\makeatletter
\DeclareRobustCommand\onedot{\futurelet\@let@token\@onedot}
\def\@onedot{\ifx\@let@token.\else.\null\fi\xspace}

\def\eg{\emph{e.g}\onedot} 
\def\ie{\emph{i.e}\onedot}

\makeatother

\theoremstyle{plain}

\theoremstyle{definition}

\theoremstyle{remark}

\usepackage[table]{xcolor}
\usepackage{colortbl}
\usepackage[textsize=tiny]{todonotes}
\usepackage{tabularx}
\usepackage{textcomp}
\usepackage{tabu} 
\usepackage{multirow}
\usepackage{makecell}

\definecolor{mycolor}{HTML}{a7caea}

\definecolor{mygreen}{HTML}{95BAA6}
\definecolor{myblue}{HTML}{4b8dbc}
\definecolor{myyellow}{HTML}{E2CD89}
\definecolor{motivationblue}{HTML}{5190BB}
\definecolor{motivationred}{HTML}{CAAAA6}
\definecolor{motivationgreen}{HTML}{5BB3A7}

\icmltitlerunning{\textsc{Light Forcing}: Accelerating Autoregressive Video Diffusion via Sparse Attention}

\begin{document}

\twocolumn[
  \icmltitle{\textsc{Light Forcing}: Accelerating Autoregressive Video Diffusion via \\ Sparse Attention}

  % It is OKAY to include author information, even for blind submissions: the
  % style file will automatically remove it for you unless you've provided
  % the [accepted] option to the icml2026 package.

  % List of affiliations: The first argument should be a (short) identifier you
  % will use later to specify author affiliations Academic affiliations
  % should list Department, University, City, Region, Country Industry
  % affiliations should list Company, City, Region, Country

  % You can specify symbols, otherwise they are numbered in order. Ideally, you
  % should not use this facility. Affiliations will be numbered in order of
  % appearance and this is the preferred way.
  \icmlsetsymbol{equal}{*}

  \begin{icmlauthorlist}
    \icmlauthor{Chengtao Lv}{ntu,comp_lab}
    \icmlauthor{Yumeng Shi}{ntu}
    \icmlauthor{Yushi Huang}{hkust}
    \icmlauthor{Ruihao Gong}{buaa,sensetime}
    \icmlauthor{Shen Ren}{comp}
    \icmlauthor{Wenya Wang}{ntu}
  \end{icmlauthorlist}

  \icmlaffiliation{ntu}{Nanyang Technological University}
  \icmlaffiliation{comp_lab}{AUMOVIO-NTU Corporate Lab}
  \icmlaffiliation{hkust}{Hong Kong University of Science and Technology}
  \icmlaffiliation{buaa}{Beihang University}
  \icmlaffiliation{sensetime}{Sensetime Research}
  \icmlaffiliation{comp}{AUMOVIO Singapore Pte Ltd}

  \icmlcorrespondingauthor{Wenya Wang}{\texttt{wangwy@ntu.edu.sg}}
  \icmlcorrespondingauthor{Ruihao Gong}{\texttt{gongruihao@buaa.edu.cn}}

  % You may provide any keywords that you find helpful for describing your
  % paper; these are used to populate the "keywords" metadata in the PDF but
  % will not be shown in the document
  \icmlkeywords{Machine Learning, ICML}

  \vskip 0.3in
]

% this must go after the closing bracket ] following \twocolumn[ ...

% This command actually creates the footnote in the first column listing the
% affiliations and the copyright notice. The command takes one argument, which
% is text to display at the start of the footnote. The \icmlEqualContribution
% command is standard text for equal contribution. Remove it (just {}) if you
% do not need this facility.

% Use ONE of the following lines. DO NOT remove the command.
% If you have no special notice, KEEP empty braces:
\printAffiliationsAndNotice{}  % no special notice (required even if empty)
% Or, if applicable, use the standard equal contribution text:
% \printAffiliationsAndNotice{\icmlEqualContribution}

\begin{abstract}
  Advanced autoregressive (AR) video generation models have improved visual fidelity and interactivity, but the quadratic complexity of attention remains a primary bottleneck for efficient deployment. While existing sparse attention solutions have shown promise on bidirectional models, we identify that applying these solutions to AR models leads to considerable performance degradation for two reasons: isolated consideration of chunk generation and insufficient utilization of past informative context. Motivated by these observations, we propose \textsc{Light Forcing}, the \textit{first} sparse attention solution tailored for AR video generation models. It incorporates a \textit{Chunk-Aware Growth} mechanism to quantitatively estimate the contribution of each chunk, which determines their sparsity allocation. This progressive sparsity increase strategy enables the current chunk to inherit prior knowledge in earlier chunks during generation. Additionally, we introduce a \textit{Hierarchical Sparse Attention} to capture informative historical and local context in a coarse-to-fine manner. Such two-level mask selection strategy (i.e., frame and block level) can adaptively handle diverse attention patterns. Extensive experiments demonstrate that our method outperforms existing sparse attention in quality (e.g., 84.5 on VBench) and efficiency (e.g., $1.2{\sim}1.3\times$ end-to-end speedup). Combined with other efficient solutions, \textsc{Light Forcing} further achieves a $2.0{\sim}3.0\times$ end-to-end speedup across diverse GPUs (e.g., 27.4\,FPS on RTX 5090 and 33.9\,FPS on H100). Code is released via this \href{https://github.com/chengtao-lv/LightForcing}{link}.

  % Combined with FP8 quantization and LightVAE, \textsc{Light Forcing} further achieves a $2.3\times$ speedup and 19.7\,FPS on an RTX~5090 GPU. Code will be released at \href{https://github.com/chengtao-lv/LightForcing}{https://github.com/chengtao-lv/LightForcing}.

\end{abstract}

\section{Introduction}
% auto

Recent notable advancements in video generation~\cite{wan2025wan,sun2024hunyuan} have revolutionized artificial intelligence-generated content (AIGC). This progress can largely be attributed to the emergence of diffusion transformers (DiT)~\cite{peebles2023scalable}, which leverage bidirectional attention to denoise all frames simultaneously. While video diffusion models (VDMs) can generate temporally consistent and long-duration videos, they struggle with temporal scalability, interactivity, and real-time deployment. In contrast, autoregressive (AR) video generation models naturally emerge as a more promising alternative, better suited to tackle these constraints. Moreover, recent AR models~\cite{yin2025slow,cui2025self} replace lossy vector quantization techniques~\cite{van2017neural} with a chunk-by-chunk generation paradigm, yielding improved visual fidelity and interactivity. This also enables real-time applications in diverse downstream tasks, such as game simulation~\cite{decart2024oasis,bruce2024genie,parker2024genie} and robot learning~\cite{yang2023learning,li2025unified}.

\begin{figure}[t!]
   \centering
        \includegraphics[width=0.44\textwidth]{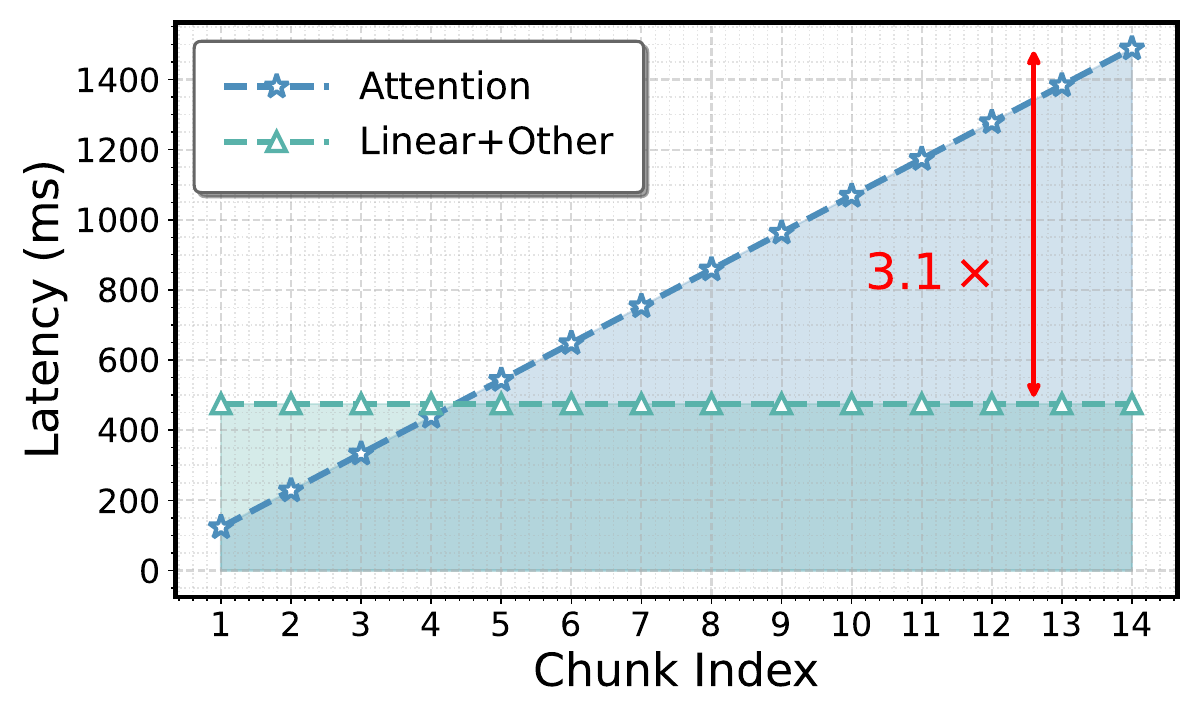}
        \caption{Runtime comparison of attention versus other components across chunk indices for Self Forcing~\cite{huang2025self} 1.3B on RTX~5090. When the chunk index reaches 14, attention accounts for approximately $\sim$75\% of the total latency.}
        \label{fig:intro}
\end{figure}

Similar to bidirectional VDMs, the quadratic computational complexity of spatiotemporal 3D full attention in AR models still remains a major bottleneck for efficient deployment. As illustrated in Fig.~\ref{fig:intro}, when generating a 480p video using Self-Forcing~\cite{huang2025self} 1.3B, the attention consumes nearly three times the runtime of all other components combined (\ie, linear layers, RoPE, \textit{etc.}) at the last chunk. To mitigate the computational costs, one simple solution is to adopt various sparse attention methods introduced for bidirectional models~\cite{zhang2025fast,xi2025sparse,yang2025sparse,zhang2025spargeattention,zhang2025sla,zhang2025vsa,li2025radial}. These approaches mainly identify critical blocks utilizing either static~\cite{zhang2025fast,li2025radial} or dynamic~\cite{wu2025vmoba,zhang2025vsa,zhang2025sla} sparse patterns in advance, and thus compute attention scores only for a small subset of tokens.

% various sparse attention methods for bidirectional models~\cite{zhang2025fast,xi2025sparse,yang2025sparse,zhang2025spargeattention,zhang2025sla,zhang2025vsa,li2025radial} have been proposed successively before. These approaches identify critical blocks utilizing either static~\cite{zhang2025fast,li2025radial} or dynamic~\cite{wu2025vmoba,zhang2025vsa,zhang2025sla} sparse patterns in advance, and thus compute attention scores only for a small subset of tokens.
% and thus avoid a large amount of attention computing.

% However, we observe that directly applying existing sparse attention solutions to autoregressive (AR) models introduces two unique challenges, which lead to significantly degraded generation quality compared to dense attention.
However, directly applying these sparse attention solutions to autoregressive (AR) models leads to significantly degraded generation quality compared to dense attention. We conduct in-depth investigations and observe that this performance drop arises from two primary aspects: \ding{192} sparse attention exacerbates the accumulation errors in AR models (\eg, over-saturation in later chunks), while prior works largely ignore the heterogeneous contributions of different chunks to the global error accumulation. Our key insight is that, during denoising, the current chunk is essentially predicting the next noise level conditioned on past clean chunks. Therefore, later chunks are naturally prone to inheriting the quality of the past chunks. \ding{193} Another insight is the insufficient utilization of past key context. For each query block, the critical historical information varies significantly across model layers, attention heads, and denoising timesteps. However,  existing methods (\eg, sliding window attention~\cite{beltagy2020longformer} or adding chunk sinks~\cite{yang2025longlive,liu2025rolling}), inevitably discard part of this information, thereby harming long-range consistency and the richness of motion in generated videos.

Motivated by these findings, we propose \textsc{Light Forcing}, an efficient variant specifically designed towards any autoregressive video generation models harnessing sparse attention. Specifically, \ding{192} we introduce a \textit{Chunk-Aware Growth} (CAG) mechanism to quantitatively estimate the contributions of each chunk. Unlike chunk-agnostic policies that treat chunk generation in isolation, we view the generation of the current chunk as a further few-step denoising process conditioned on the previous clean chunk. From a theoretical perspective, we formulate the final sparsity allocation for each chunk as determined by its global accumulation error, which depends on two components (\ie, the corresponding denoising steps and the score estimation). In other words, our method allocates lower sparsity priorities to earlier chunks, and progressively increases the sparsity in later chunks as they can inherit the structured knowledge stored in earlier chunks. \ding{193} We propose \textit{Hierarchical Sparse Attention} (HSA), which preserves both global and local perception ability under a fixed computational budget. Specifically, HSA adopts a coarse-to-fine pipeline that selects sparse masks at both the frame and block levels for each query block, enabling flexible and versatile attention modeling. This two-level strategy efficiently captures informative historical context while maintaining fast execution, thereby achieving an effective trade-off between model performance and computational cost.

We conduct extensive experiments to evaluate the effectiveness of our \textsc{Light Forcing}. We compare our method with state-of-the-art sparse attention approaches on three autoregressive video generation models: Self Forcing~\cite{huang2025self}, LongLive~\cite{yang2025longlive}, and Infinite-Forcing~\cite{infinite-forcing}. We report results on two benchmarks, VBench~\cite{huang2024vbench} and VBench-Long~\cite{huang2025vbench++}. The results show that our method consistently outperforms existing approaches in both generation quality and latency, and even surpasses dense attention in several metrics. For example, on Self Forcing~\cite{huang2025self}, our method achieves a total score of 84.5 while providing 1.3$\times$ end-to-end and 3.79$\times$ attention speedup. Furthermore, we provide complementary acceleration techniques, including FP8 linear layers, kernel fusion (\ie, RoPE, RMSNorm, \textit{etc.}), and an efficient VAE (\ie, LightVAE~\cite{lightx2v}). With plug-and-play configuration files, these optimizations enable approximately $2.0{\sim}3.0\times$ end-to-end speedup across diverse GPUs (\eg, 27.4\,FPS, 16.8\,FPS, and 33.9\,FPS on a single RTX~5090, A100, and H100 GPU, respectively).

% For example, on Self Forcing~\cite{huang2025self}, our method achieves a total score of 84.5 while providing 1.3$\times$ end-to-end and 3.3$\times$ attention speedup. Furthermore, when combined with FP8 quantization and LightVAE~\cite{lightx2v}, \textsc{Light Forcing} reaches 19.7 FPS, enabling real-time video generation on a consumer-grade GPU (RTX 5090) for the first time.

To summarize, our main contributions are threefold:
\begin{itemize}[leftmargin=*, nosep]
    \item To the best of our knowledge, \textsc{Light Forcing} is the first sparse attention solution specifically designed for autoregressive video generation models.
    \item We present \textit{Chunk-Aware Growth (CAG).} We allocate higher attention budgets to earlier chunks and progressively decay for later chunks, effectively reducing error propagation while preserving efficiency.
    \item We propose \textit{Hierarchical Sparse Attention (HSA)}, which captures global and local dependencies via coarse-to-fine frame and block selection.
    \item Extensive experiments demonstrate the superior performance (\eg, 84.5 on VBench) and real-time generation (\eg, $2.0{\sim}3.0\times$ end-to-end speedup) of \textsc{Light Forcing} across diverse GPUs.
\end{itemize}
% Extensive experiments are

% first retrieves a set of key frames and then performs dynamic sparse attention over the selected frames.

% This combination of frame-level and block-level sparsity enables fine-grained coverage of informative historical context for each query block, exhibiting strong flexibility and versatility.

% based on which we assign different sparsity levels to different chunks.

% efficiency bottleneck

% error accumulation --> inter-chunk sparse attention makes it worse, --> 
% intra-chunk

\section{Related Work}
\subsection{Autoregressive Video Diffusion}
Compared with bidirectional video diffusion models~\cite{wan2025wan,yang2024cogvideox,sun2024hunyuan} that denoise all frames jointly, autoregressive video generation models~\cite{zhang2026frame,huang2025self,gu2025long,kodaira2025streamdit,henschel2025streamingt2v} generate the next token or frame sequentially, and are thus inherently more suitable for real-time streaming applications. Early approaches~\cite{hu2024acdit,gao2024ca2} adopted Teacher Forcing (TF), where training is conditioned on ground-truth tokens, but they suffer from reduced visual fidelity when generating long videos. Conversely, Diffusion Forcing~\cite{chen2024diffusion} is trained with conditioning at arbitrary noise levels and has been adopted in models such as SkyReels-V2~\cite{chen2025skyreels} and Magi-1~\cite{teng2025magi}. CausVid~\cite{yin2025slow} employs block causal attention, distilling a bidirectional teacher
to a few-step causal student via distribution matching distillation~\cite{yin2024improved}. More recently, Self Forcing~\cite{huang2025self} introduced a novel post-training paradigm that mitigates error accumulation arising from train-test misalignment. Subsequent works, including Rolling Forcing~\cite{liu2025rolling}, LongLive~\cite{yang2025longlive}, Self Forcing++~\cite{cui2025self}, and Reward Forcing~\cite{lu2025reward}, further address the limitation on the achievable generation length, object/scene dynamics or color drifts. Nevertheless, although autoregressive models with only a few denoising steps (\eg, 4 steps) have substantially reduced latency, real-time generation on resource-constrained devices still remains challenging.

\subsection{Sparse Attention}
A large body of work~\cite{wu2025vmoba,xu2025xattention,xi2025sparse,yang2025sparse,zhang2025spargeattention,wu2025pack,dalal2025one} has explored how to alleviate the runtime bottleneck caused by quadratic-complexity attention in bidirectional video diffusion models, covering low-bit attention~\cite{zhang2024sageattention,zhang2024sageattention2} and linear attention~\cite{xie2024sana,chen2025sana,huang2025linvideo}. Another promising line of work focuses on sparse attention, where approaches can be roughly categorized by whether they follow \textit{static} or \textit{dynamic} patterns to identify critical tokens with block-wise granularity. Static schemes~\cite{zhang2025fast,li2025radial,hassani2025generalized} usually prescribe sparsity masks via handcrafted patterns, such as neighborhood~\cite{hassani2025generalized,zhang2025fast} or spatiotemporal structures~\cite{xi2025sparse}. In contrast, dynamic solutions~\cite{zhang2025spargeattention,zhang2025sla,wu2025vmoba,cai2025mixture} additionally introduce an online identification stage. These methods either utilize 1D~\cite{zhang2025spargeattention,wu2025vmoba,zhang2025sla} or 3D~\cite{zhang2025vsa,wu2025vmoba} mean pooling to aggregate blocks, and estimate their importance subsequently. Clustering-based strategies~\cite{yang2025sparse} instead group semantically similar tokens together. In addition, several emerging ~\textit{hybrid} attention mechanisms have been applied to video generation, including mixtures across different attention types (\eg, combining linear attention and softmax attention~\cite{zhang2025sla}) and across different sparsity levels (\eg, gating of twin-level~\cite{zhang2025vsa} or pyramid-level sparse representation~\cite {li2025psa,zhou2025trainable}). However, the exploration of sparse attention for autoregressive video generation remains largely uncharted.

% Among them, 1D mean-pooling–based criteria are the most widely used, and have recently been extended to 3D in video generation.

\section{Preliminaries}
\noindent\textbf{Autoregressive video diffusion modeling.} Autoregressive~(AR) video diffusion models decompose video synthesis into \emph{inter-chunk} autoregression and \emph{intra-chunk} diffusion, combining the chain-rule factorization for temporal dependency modeling with the expressive denoising capability of diffusion models for high-fidelity frame generation. Specifically, given condition $c$, the joint distribution of an $N$-frame video sequence $\boldsymbol{x}^{1:N}$ is expressed as
\begin{equation}
p_{\theta}(\boldsymbol{x}^{1:N}|c) = \prod_{i=1}^N p_{\theta}(\boldsymbol{x}^i \mid \boldsymbol{x}^{<i}, c).
\end{equation}
This formulation generates frames sequentially, where each conditional term $p_{\theta}(\boldsymbol{x}^i \mid \boldsymbol{x}^{<i}, c)$ is approximated by a few-step diffusion generator conditioned on KV cache (\ie, previous clean frames)~\cite{huang2025self,cui2025self,yang2025longlive}. Specifically, the conditional term $p_\theta(\boldsymbol{x}^i \mid \boldsymbol{x}^{<i}, c)$ can be defined as $f_{\theta,t_1}\circ f_{\theta,t_2}\circ \cdots \circ f_{\theta,t_T}\!\left(\boldsymbol{x}^i_{t_T}\right)$, where $\boldsymbol{x}^i_{t_T}\sim\mathcal{N}(\mathbf{0},\mathbf{I})$ and each transition is given by
\begin{equation}
    f_{\theta,t_j}\!\left(\boldsymbol{x}^i_{t_j}\right)
=
\Psi\!\Big(
G_\theta(\boldsymbol{x}^i_{t_j},\, t_j,\, \boldsymbol{x}^{<i},\, c),
\, \boldsymbol{\epsilon}_{t_{j-1}},\, t_{j-1}
\Big),
\end{equation}
where $G_\theta(\boldsymbol{x}^i_{t_j},\, t_j,\, \boldsymbol{x}^{<i},\, c)$ corresponds to the denoised estimate $\hat{\boldsymbol{x}}^{\,i}_0$, \ie, a prediction of the clean chunk $i$ from the current noisy state $\boldsymbol{x}^i_{t_j}$ under the autoregressive context $\boldsymbol{x}^{<i}$ and condition $c$. The operator $\Psi(\cdot)$ denotes the forward corruption (re-noising) mapping that injects Gaussian noise at a lower noise level to produce the next state $\boldsymbol{x}^{i}_{t_{j-1}}$ for subsequent denoising. Advanced few-step AR video diffusion models often adopt the probability flow ODE formulation to define the forward noising trajectory and inject Gaussian noise~\cite{song2023consistency}, \ie,
$(1-\sigma_{t_{j-1}})\hat{\boldsymbol{x}}^{\,i}_0+\sigma_{t_{j-1}}\boldsymbol{\epsilon}_{t_{j-1}}$,
where $\boldsymbol{\epsilon}_{t_{j-1}}\sim\mathcal{N}(\mathbf{0},\mathbf{I})$ and $\sigma_{t_{j-1}}$ controls the noise level.

\begin{figure*}[t!]
   \centering
        \includegraphics[width=0.9\textwidth]{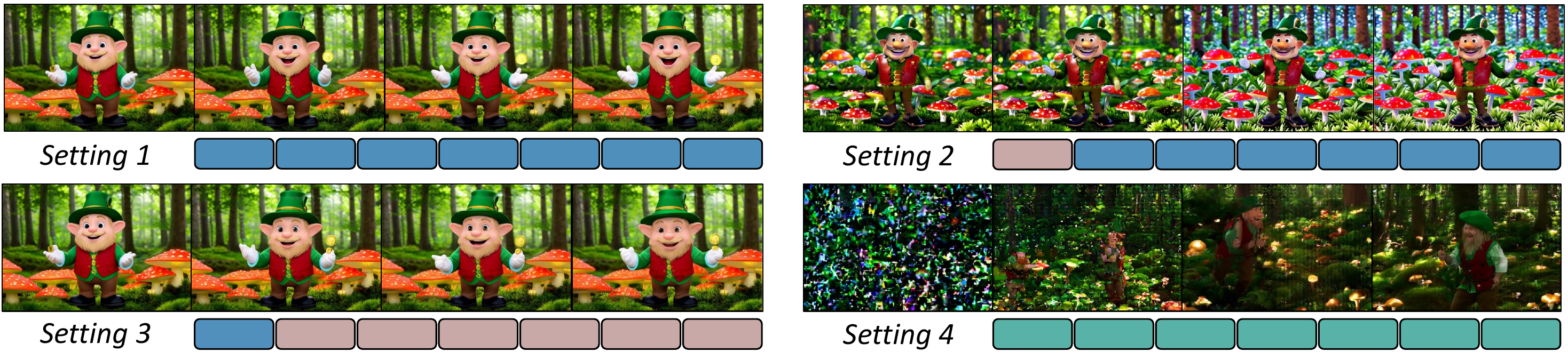}
        \caption{Comparison of different visual generation examples (\ie, 7 chunks for 21 latent frames), where \textcolor{motivationblue}{blue}, \textcolor{motivationred}{red}, and \textcolor{motivationgreen}{green} boxes denote attention sparsity rates of 0\%, 80\%, and 90\%, respectively.}
        \label{fig:method1_analysis}
\end{figure*}

\noindent\textbf{Blockwise sparse attention.}\label{sec:pre_sparse_attention} Practical sparse-attention systems enforce sparsity at the \emph{block} (tile) granularity to better match modern accelerator hardware, enabling high utilization and efficient memory access patterns in GPU kernels such as FlashAttention~\cite{dao2023flashattention}. Concretely, given $\vq,\vk,\vv\in\mathbb{R}^{n\times d}$, we partition\footnote{For simplicity, we assume $\vq$ and $\vk/\vv$ have the same shape.} the sequence dimension into blocks and form
\[
\vq=\big[\vq_1;\ldots;\vq_{n_q}\big],
\vk=\big[\vk_1;\ldots;\vk_{n_k}\big],
\vv=\big[\vv_1;\ldots;\vv_{n_k}\big],
\]
where $\vq_i\in\mathbb{R}^{b_q\times d}$ and $\vk_j,\vv_j\in\mathbb{R}^{b_{kv}\times d}$, with $n_q=\lceil n/b_q\rceil$ and $n_k=\lceil n/b_{kv}\rceil$. We further define a \emph{block mask} $\mB\in\{0,1\}^{n_q\times n_k}$, where $\mB_{ij}=1$ indicates that the $(i,j)$ tile is active. Block-sparse attention can then be written as
\begin{equation}
\mathrm{SparseAttn}(\vq,\vk,\vv;\mB)
=\mathrm{softmax}\!\left(\frac{\vq\vk^\top}{\sqrt{d_k}} \odot \mM(\mB)\right)\vv,
\label{eq:block_sparse_attn}
\end{equation}
where $\mM(\mB)\in\{0,1\}^{n\times n}$ expands $\mB$ to an element-wise mask that is constant within each $(b_q\times b_{kv})$ tile, and $\odot$ denotes element-wise multiplication. Importantly, efficient implementations do not materialize $\mM(\mB)$. Instead, they compute only the tile products $\vq_i\vk_j^\top$ and the corresponding value aggregation for indices $(i,j)$ with $\mB_{ij}=1$, skipping entire tiles when $\mB_{ij}=0$. Consequently, the computational and memory costs scale with the number of \emph{active} blocks rather than $n^2$, while retaining GPU-friendly dense computation within each tile.

% \begin{equation}
% \begin{aligned}
% \boldsymbol{x}^{i}_{t_{j-1}}
% &=\Psi\!\left(\hat{\boldsymbol{x}}^{\,i}_0,\,\boldsymbol{\epsilon}_{t_{j-1}},\,t_{j-1}\right) \\
% &=\left(1-\sigma_{t_{j-1}}\right)\hat{\boldsymbol{x}}^{\,i}_0+\sigma_{t_{j-1}}\boldsymbol{\epsilon}_{t_{j-1}},
% \boldsymbol{\epsilon}_{t_{j-1}}\sim\mathcal{N}(\mathbf{0},\mathbf{I}).
% \end{aligned}
% \end{equation}

% \begin{equation}
% \boldsymbol{x}^{i}_{t_{j-1}}
% =
% \Psi\!\left(\boldsymbol{x}^{0},\, \boldsymbol{\epsilon}_{t_{j-1}},\, t_{j-1}\right)
% =
% \left(1-\sigma_{t_{j-1}}\right)\boldsymbol{x}^{0}
% +
% \sigma_{t_{j-1}}\boldsymbol{\epsilon}_{t_{j-1}},
% \qquad
% \boldsymbol{\epsilon}_{t_{j-1}}\sim\mathcal{N}(\mathbf{0},\mathbf{I}).
% \end{equation}

\section{\textsc{Light Forcing}}

\subsection{Chunk-Aware Growth Mechanism}
\label{sec:method1}
Many acceleration techniques for bidirectional video diffusion models, including feature caching~\cite{huang2024harmonica,liu2025timestep,ma2024learning} and sparse attention~\cite{li2025radial,zhang2025fast}, have observed pronounced sensitivity across different timesteps and layers. However, directly applying these \emph{chunk-agnostic} policies of bidirectional models to few-step autoregressive video diffusion can be problematic: they ignore the heterogeneous contribution of different chunks to the \emph{global accumulation error} that compounds over autoregressive rollout, and thus can easily trigger severe quality degradation or even collapse. To build intuition, we conduct several simple toy experiments that visually illustrate how generation behavior varies across chunks (as shown in Fig.~\ref{fig:method1_analysis}).

% First, we apply a moderately sparse attention ratio (\eg, 80\%) either to the first chunk (Setting~2) or to the subsequent chunks 2-7 (Setting~3). Surprisingly, we observe an interesting phenomenon: Setting~2 incurs an irreversible loss of visual quality (even over-saturation in the later chunks) that cannot be recovered even if later chunks revert to dense attention. Later frames in Setting~2 exhibit severe over-saturation and exposure-bias artifacts. In contrast, Setting~3 achieves generation quality that is nearly lossless from Setting~1 even when only the first chunk is kept in dense attention. This further implies that once satisfactory priors are established in the first (or other early) chunk(s), subsequent chunks can readily inherit and propagate these priors with little difficulty. Intuitively, earlier chunks should adopt lower sparsity, while later chunks can tolerate higher sparsity. Therefore, a natural question arises: \textit{Can we quantitatively allocate the sparsity budget across chunks?}

First, we apply a moderately sparse attention ratio (\eg, 80\%) either to the first chunk (Setting~2) or to the subsequent chunks 2-7 (Setting~3). Surprisingly, we observe an interesting phenomenon: Setting~2 incurs an irreversible loss of visual quality (even over-saturation in the later chunks) that cannot be recovered even if later chunks revert to dense attention. Later frames in Setting~2 exhibit severe over-saturation and exposure-bias artifacts. In contrast, Setting~3 achieves generation quality that is nearly lossless from Setting~1 even when only the first chunk is kept in dense attention. This further implies two observations: \ding{192} the first chunk acts as a \textit{visual anchor} for the entire autoregressive rollout; \ding{193} once satisfactory priors are established in the first (or other early) chunk(s), subsequent chunks can readily inherit and propagate these priors with little difficulty. Therefore, we keep dense attention for the first chunk and allocate lower sparsity to other early chunks, while later chunks can tolerate higher sparsity. A natural question then arises: \textit{Can we quantitatively allocate the sparsity budget across chunks?}

\begin{figure*}[t!]
   \centering
        \includegraphics[width=0.9\textwidth]{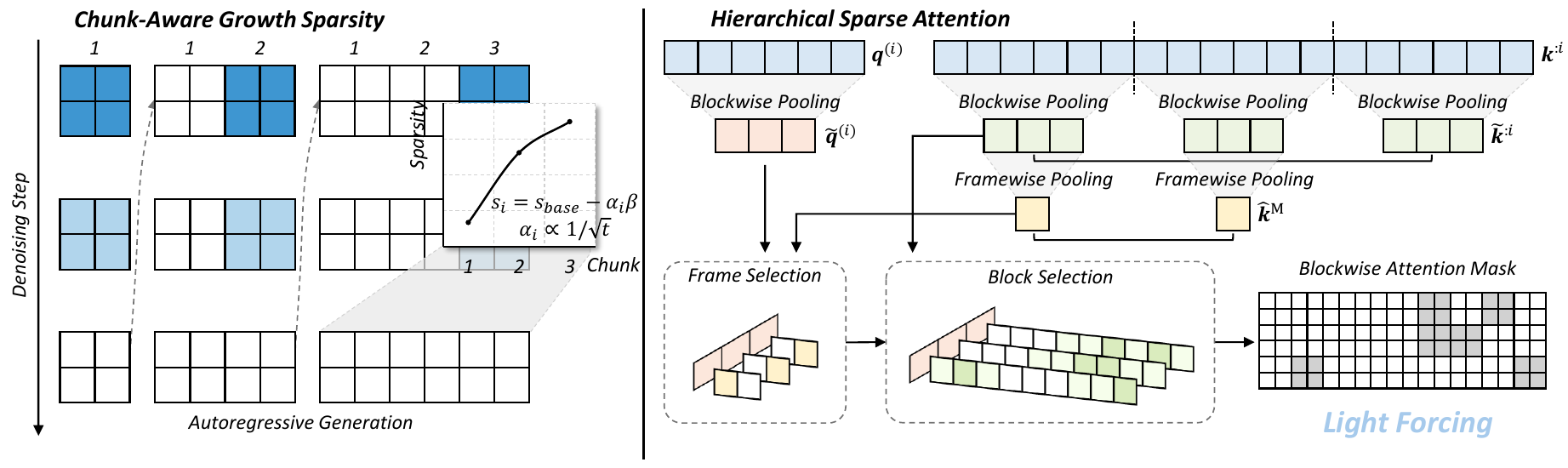}
        \caption{Overview of \textsc{Light Forcing}. The left subfigure illustrates our \textit{ Chunk-Aware Growth} (Sec.~\ref{sec:method1}) strategy for sparsity allocation across different chunks. The right subfigure demonstrates how \textit{Hierarchical Sparse Attention} (Sec.~\ref{sec:method2}) is utilized to efficiently retrieve long-range historical context. Note that a chunk corresponds to a group of frames processed in a single generation (\eg, 3 frames in practice). For simplicity, we visualize each chunk as a single frame in the overview.}
        \label{fig:overview}
\end{figure*}

\textbf{Sparsity-Induced Error.} To solve the problem, we further explore generation performance under a radical sparsity rate (\ie, 90\%) in Setting 4, and observe that as time progresses, later chunks gradually become clearer compared to the initially Gaussian-noise-like appearance, suggesting that $G_\theta(\boldsymbol{x}^i_{t_j},\, t_j,\, \boldsymbol{x}^{<i},\, c)$ performs denoising toward the next noise level compared with $\boldsymbol{x}^{i-1}$. Moreover, we posit that $G_\theta(\boldsymbol{x}^i_{t_j},\, t_j,\, \boldsymbol{x}^{<i},\, c)$ essentially continues denoising for $T$ additional steps starting from the noisy level of $\boldsymbol{x}^{i-1}$ (verified in the Appendix). 

These analyses reveal that sparsity effects can be reflected by the noise level in the final generated clean chunk $\boldsymbol{x}^{i}_{t_0}$ ($i=1,\ldots, N$). To capture this, we measure the sparsity-induced error of chunk $i$ as the variation distance $\mathrm{TV}(\cdot,\cdot)$ between the clean data distribution $p$ and the generated distribution of $\boldsymbol{x}^{i}_{t_0}$, denoted $q_t$, with its noise level:
\begin{align}
\mathrm{TV}(q_t, p)
&\le C_1 \frac{d^2 \log^3 T}{\sqrt{T}}
+ C_2 \sqrt{d}\,\varepsilon_{\text{score}} \log^2 T,
% &\le \sqrt{\tfrac{1}{2}\mathrm{KL}(q_1 \| p_1)} \\
\label{eq:bound}
\end{align}
where $C_1$, $C_2$, $\log^2 T$, $\log^3 T$ are constants\footnote{Logarithmic factors of the form $\log^k T$ arise only as technical amplification constants in the proof and do not affect the asymptotic complexity.} do not affect the asymptotic complexity. From this inequality, we can interpret the first term as the finite-step sampling error, which is $\propto 1/\sqrt{T}$, while the second term captures the effect of score estimation error, reflecting the approximation error induced by imperfect model learning.

% The above analysis reveals that different chunks contribute unequally to global accumulation error and the final generation quality. To capture this effect, we quantify each chunk's generation difficulty using distances measured at different denoising steps.

% Motivated by the above analysis, we quantify each chunk's generation difficulty and its contribution to global accumulation error using the distance at different denoising steps. Following~\cite{li2023towards}, we have
% \begin{align}
% \mathrm{TV}(q_1, p_1)
% &\le C_1 \frac{d^2 \log^3 T}{\sqrt{T}}
% + C_1 \sqrt{d}\,\varepsilon_{\text{score}} \log^2 T
% % &\le \sqrt{\tfrac{1}{2}\mathrm{KL}(q_1 \| p_1)} \\
% \end{align}
% where $\mathrm{TV}(q_1,p_1)$ denotes the total variation distance between the true distribution $p_1$ and the generated distribution $q_1$, and $C_1$ is a universal constant. The $\log^2 T$ and $\log^3 T$ terms are logarithmic factors introduced by loosening the bound. From this expression, we can interpret the first term as the finite-step sampling error, which is $\propto 1/\sqrt{T}$, while the second term captures the effect of score estimation error, reflecting the approximation error induced by imperfect model learning. 

% \textbf{Sparsity Allocation.} Intuitively, we should lower the sparsity ratio for chunks with more errors to preserve generation quality. Leveraging this insight, we propose a \textit{Chunk-Aware Growth} (CAG) strategy that considers both the finite-step sampling error (Term~1) and the score estimation error (Term~2). For chunk $i$, the sparsity ratio $s_i$ can be written as
\textbf{Sparsity Allocation.} Intuitively, we should lower the sparsity ratio for chunks with more errors to preserve generation quality. Leveraging this insight, we propose a \textit{Chunk-Aware Growth} (CAG) strategy that considers both the finite-step sampling error (Term~1) and the score estimation error (Term~2). In practice, we set dense attention for the first chunk to preserve the initial visual anchor, and apply CAG to the remaining chunks. For chunk $i>1$, the sparsity ratio $s_i$ can be written as
\begin{equation}
s_i = s_{base} - \alpha_i \beta    
\end{equation}
where $\alpha_i$ denotes the noise level reached by the $i$-th chunk and scales as $\propto 1/\sqrt{T}$. The hyperparameter $s_{base}$ is a predefined constant that reflects the score estimation error. To solve for the modulated sparsity factor $\beta$, we enforce the total FLOPs after chunk-wise modulation to be equal to the FLOPs specified by the target sparsity ratio:
\begin{equation}
(1-s_{target})
\sum_{i=2}^{n}  l_i^q l_i^{k} d
=
\sum_{i=2}^{n} \bigl(1 -s_{base}+\alpha_i \beta )\,  l_i^q l_i^{k} d,
\end{equation}
where $s_{target}$ denotes the target sparsity ratio, and $l_i^{q}$ and $l_i^{k}$ denote the query and key sequence lengths for chunk $i$, respectively\footnote{For chunk $i$, given $\vq_i\in\mathbb{R}^{l_i^{q}\times d}$ and $\vk_i\in\mathbb{R}^{l_i^{k}\times d}$, the dominant FLOPs of attention are $\mathcal{O}(l_i^{q}l_i^{k}d)$.}. This equality yields $\beta$ and consequently determines the chunk-wise sparsity ratio $s_i$.

% To compute the modulated sparsity factor $\beta$, we have
% \begin{equation}
% (1-s_{target})
% \sum_{i=2}^{n}  l_i^q l_i^{k} d
% =
% \sum_{i=2}^{n} \bigl(1 -s_{base}+\alpha_i \beta )\,  l_i^q l_i^{k} d,
% \end{equation}
% where $s_{target}$ denotes the target sparsity ratio, and $l_i^{q}$ and $l_i^{k}$ denote the query and key sequence lengths for chunk $i$, respectively. By enforcing equal FLOPs on both sides of the equation, we can solve for $\beta$ and thus obtain $s_i$.

\subsection{Hierarchical Sparse Attention}
\label{sec:method2}

\begin{figure*}[ht]
    \centering
    \begin{minipage}{0.24\textwidth}
        \centering
        \includegraphics[width=\linewidth]{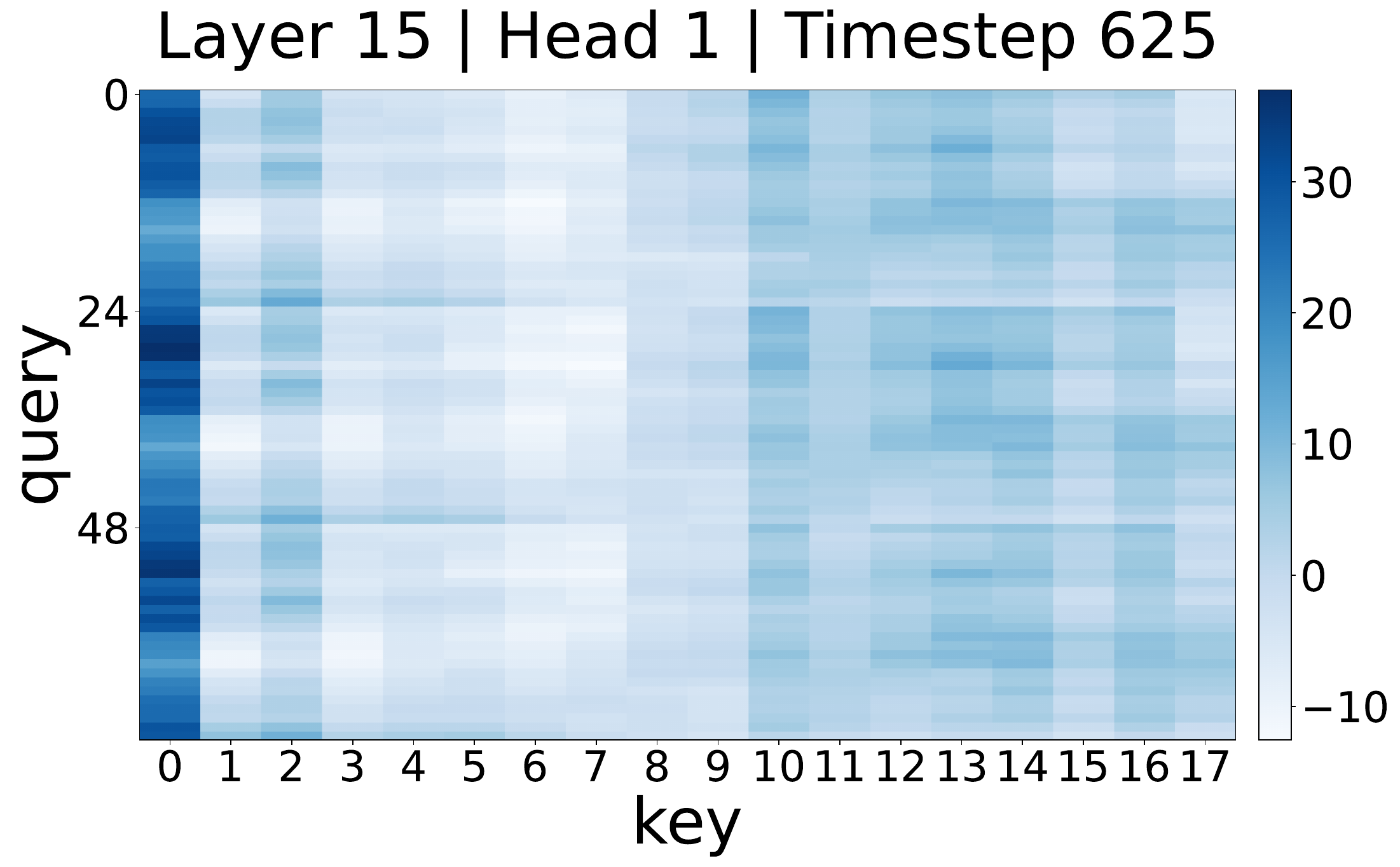}
    \end{minipage}
    \begin{minipage}{0.24\textwidth}
        \centering
        \includegraphics[width=\linewidth]{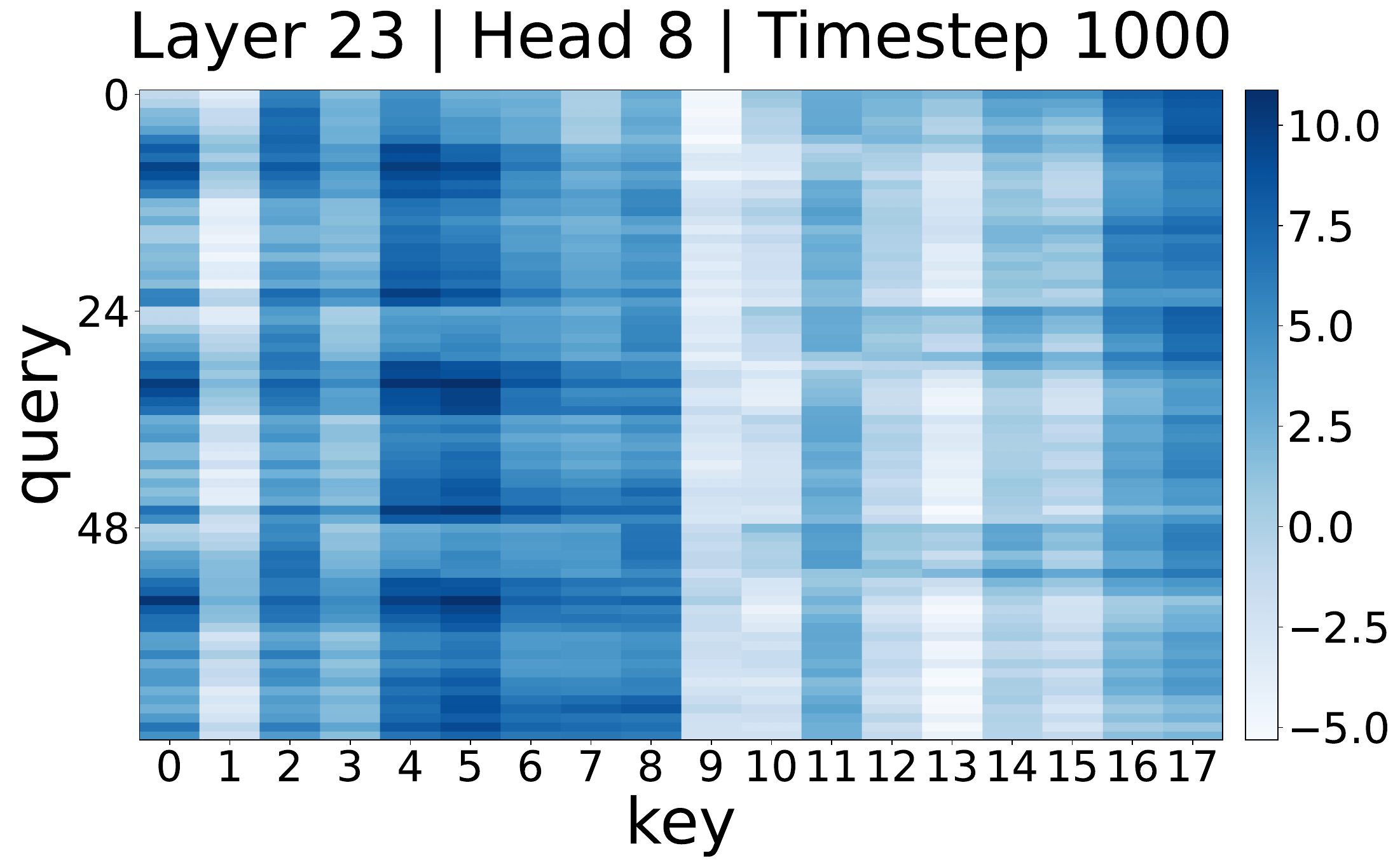}
    \end{minipage}
    \begin{minipage}{0.24\textwidth}
        \centering
        \includegraphics[width=\linewidth]{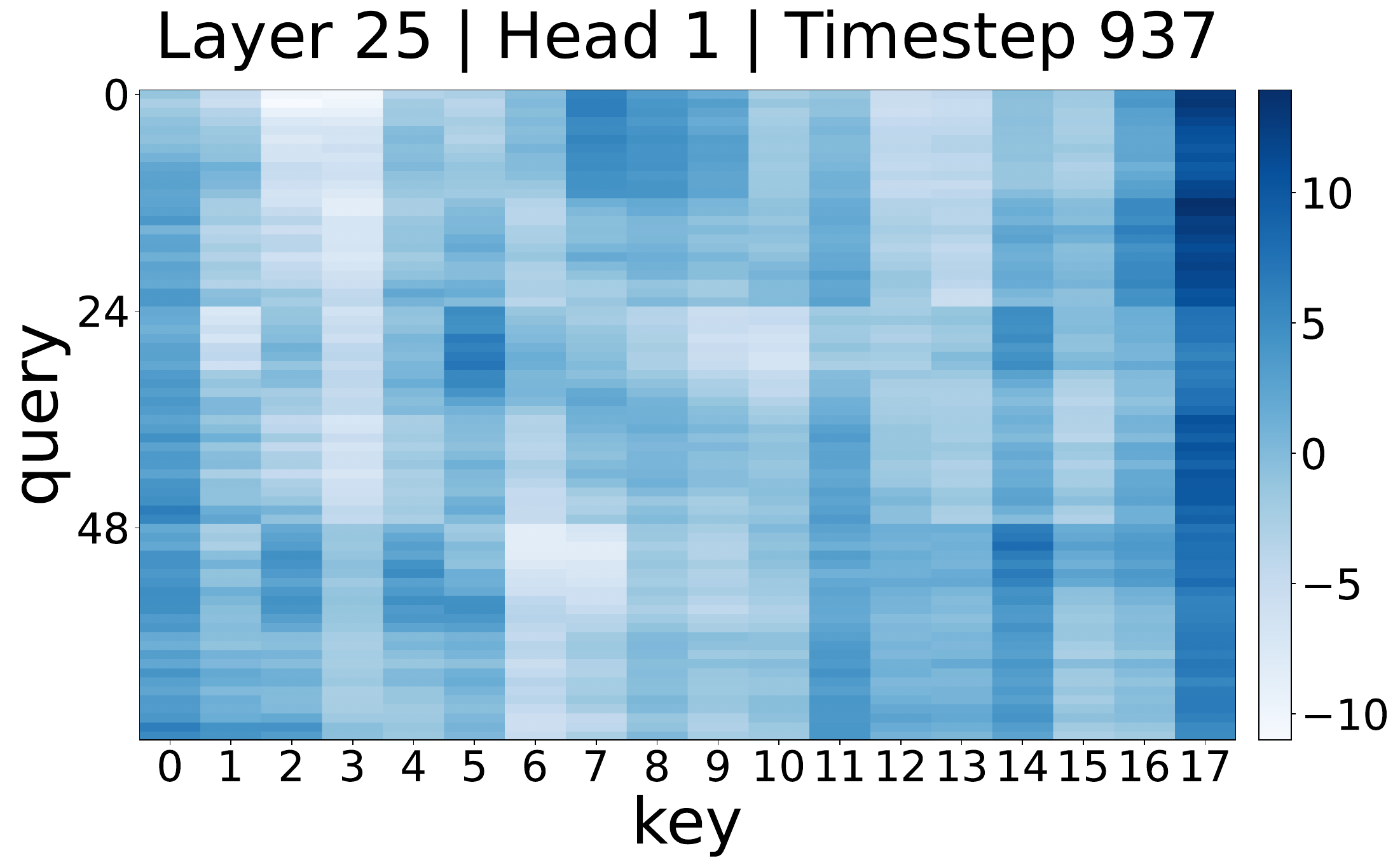}
    \end{minipage}
    \begin{minipage}{0.24\textwidth}
        \centering
        \includegraphics[width=\linewidth]{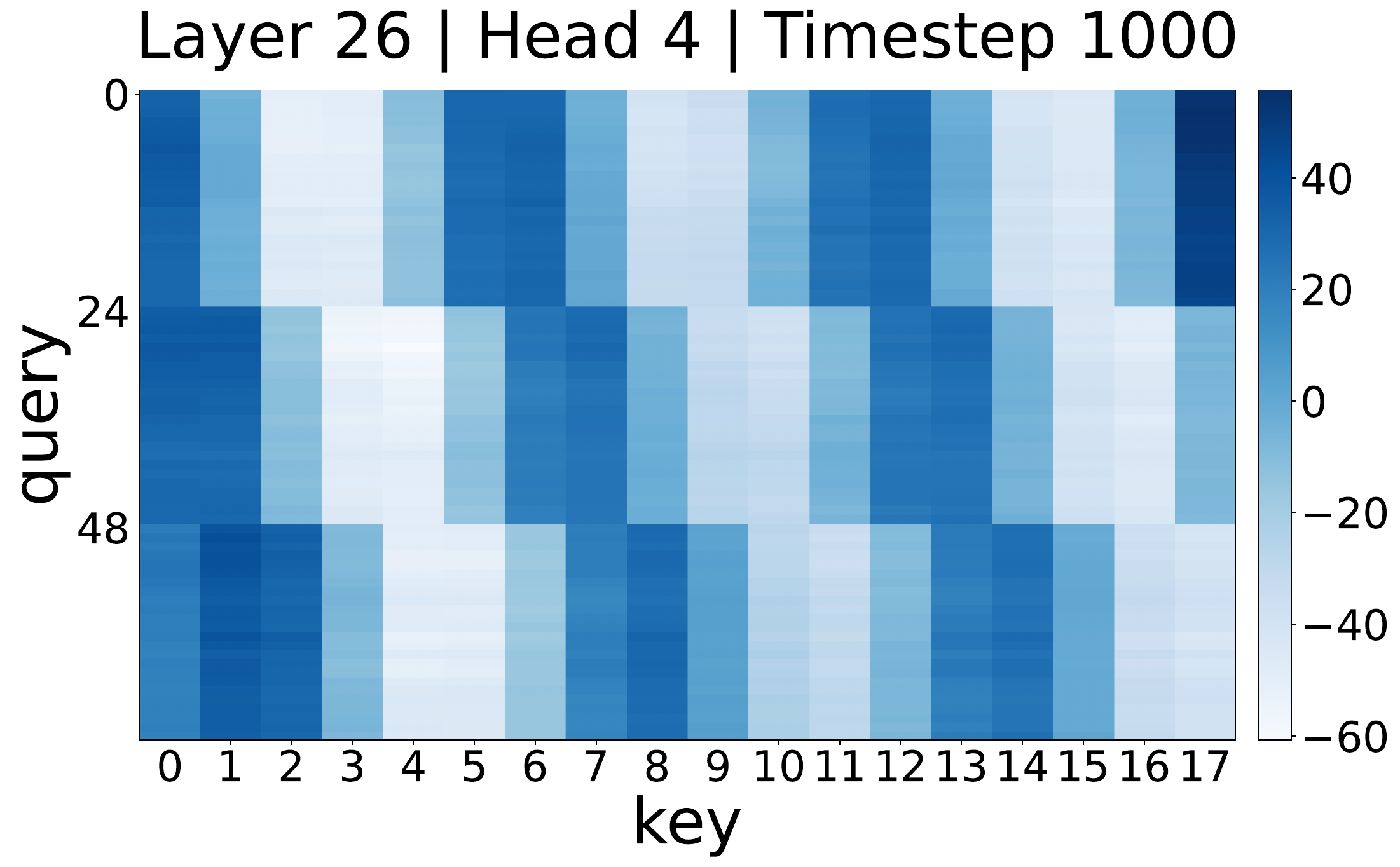}
    \end{minipage}
    \caption{Visualization of attention logits between query blocks at chunk 7 (\ie, frame 18-20, 24 blocks per frame) and all past key frames (\ie, frame 0-17) on Self Forcing~\cite{huang2025self}.}
    \label{fig:attention_map}
\end{figure*}

Another key challenge of autoregressive video generation models is that the number of historical frames grows linearly over time, making attention increasingly time-consuming and slowing down later-chunk generation. Many approaches~\cite{liu2025rolling,lu2025reward,yang2025longlive} mitigate this by adopting sliding-window attention~\cite{beltagy2020longformer} that truncates past frames to a fixed context length. While this alleviates the latency growth, it can induce history forgetting, leading to poor long-range consistency and repetitive motions in subsequent chunks. We believe that treating nearby frames as keyframes is suboptimal, since the historical frames that the current query is interested in vary across different layers, heads, and timesteps. As illustrated in Fig.~\ref{fig:attention_map}, attention between historical frames exhibits complex patterns such as diagonal, attention-sink structures, which makes it difficult for a sliding window scheme to cover all informative context.

Inspired by these findings, we propose the \textit{Hierarchical Sparse Attention} (HSA), which follows a coarse-to-fine paradigm for sparse attention on autoregressive video generation models. Specifically, each query block first retrieves a set of keyframes and then performs dynamic sparse attention over the selected frames. This dual-stage strategy not only bounds attention to a fixed computational complexity but also mitigates long-video consistency degradation and history-forgetting issues.
% (as detailed in Sec.~\ref{sec:pre_sparse_attention}) 

Formally, the process of generating chunk $i$ can be understood as computing attention between the query $\vq^{(i)}$ and the key-value pairs $\{\vk^{:i}, \vv^{:i}\}$. Here, $\vq^{(i)} \in \mathbb{R}^{(f \times n) \times d}$ and $\{\vk^{:i}, \vv^{:i}\} \in \mathbb{R}^{(i \times f \times n) \times d}$, where $f$ and $n$ denote the number of frames per chunk and the number of tokens per frame, respectively. To avoid computing attention between all key-value pairs, our HSA mainly consists of three components, \ie, Token Compression, Mask Selection, and Blockwise Sparse Attention.

\noindent\textbf{Token Compression.} We first compress $\vq^{(i)}$ at the blockwise granularity and  $\vk^{:i}$ at the blockwise and framewise granularity, which can be written as
\begin{equation}
\begin{aligned}
\text{blockwise:}\quad
&\tilde{\vq}^{(i)}=\phi\!\left(\vq^{(i)},\, b_q\right),
\tilde{\vk}^{:i}=\phi\!\left(\vk^{:i},\, b_{kv}\right),\\
\text{framewise:}\quad
&\hat{\vk}^{\mathcal{M}}
=
\phi\!\left(\tilde{\vk}^{\mathcal{M}},\, \lceil n/b_{kv}\rceil\right),
\end{aligned}
\label{eq:token_compress}
\end{equation}
where $\phi(\mathbf{x},b)$ denotes a mean pooling operator that aggregates sequential tokens with size $b$. After applying this operation, the final compressed representations have shapes $\tilde{\vq}^{(i)} \in \mathbb{R}^{(f \times \lceil n/b_{q}\rceil) \times d}$, $\tilde{\vk}^{:i} \in \mathbb{R}^{(i \times f \times \lceil n/b_{kv}\rceil) \times d}$, and $\hat{\vk}^{\mathcal{M}} \in \mathbb{R}^{m \times d}$. It is worth noting that $\hat{\vk}^{\mathcal{M}}$ does not cover all past frames\footnote{$m=(i-1)\times f-n_{sink}-n_{win}$, where $n_{sink}$ denotes the number of earliest sink frames and $n_{win}$ denotes the small number of frames in the sliding window.}, since the sink frames, the nearest frames within the sliding window, and the key-value frames within \textit{current} chunk $i$ are always selected.

\noindent\textbf{Mask Selection.}
Based on the compressed representations, we perform a hierarchical mask selection in a coarse-to-fine manner. Specifically, for each query block in chunk $i$, we first retrieve a small set of relevant historical frames using the framewise-compressed keys, and then select critical blocks within the retrieved frames using blockwise-compressed keys.

% Let $\tilde{\vq}^{(i)}_r \in \mathbb{R}^{d}$ denote the $r$-th blockwise query summary in the current chunk, where $r \in \{1,\ldots,f \times \lceil n/b_q\rceil\}$. We compute frame-level relevance scores between the query block and each past frame using the framewise-compressed keys $\hat{\vk}^{\mathcal{M}}$ as
% \begin{equation}
% p^{(i)}_r
% =
% \big\langle \tilde{\vq}^{(i)}_r,\; \hat{\vk}^{\mathcal{M}} \big\rangle
% \;\in\;
% \mathbb{R}^{m},
% \label{eq:frame_level_score}
% \end{equation}
% where $p^{(i)}_r$ denotes the vector of frame-level logits, whose entries are given by the inner products between the query block summary $\tilde{\vq}^{(i)}_r$ and each framewise-compressed key in $\hat{\vk}^{\mathcal{M}}$. We then select the most relevant past frames:

Let $\tilde{\vq}^{(i)}_r \in \mathbb{R}^{d}$ denote the $r$-th blockwise query summary in the current chunk, where $r \in \{1,\ldots,f \times \lceil n/b_q\rceil\}$. We compute frame-level relevance scores between the query block and each candidate past key in $\mathcal{M}$ using the framewise-compressed keys $\hat{\vk}^{\mathcal{M}}$ as
\begin{equation}
p^{(i)}_r
=
\big\langle \tilde{\vq}^{(i)}_r,\; \hat{\vk}^{\mathcal{M}} \big\rangle
\;\in\;
\mathbb{R}^{m},
\label{eq:frame_level_score}
\end{equation}
where $p^{(i)}_r$ denotes the vector of frame-level logits, whose entries are given by the inner products between the query block summary $\tilde{\vq}^{(i)}_r$ and each framewise-compressed key in $\hat{\vk}^{\mathcal{M}}$. We then select the most relevant candidate past keys from $\mathcal{M}$:

\begin{equation}
\mathcal{T}_r
=
\mathrm{TopK_{idx}}\!\left(p^{(i)}_r\right)
\;\cup\;
\mathcal{F}^{(i)},
\label{eq:frame_selection}
\end{equation}

where $\mathrm{TopK_{idx}}$ returns indices of the most relevant top-$k$ candidate frames and $\mathcal{F}^{(i)}$ denotes the always-selected frame set discussed above, including a few sink frames, nearest frames, and the frames within the current chunk $i$, ensuring motion smoothness and full visibility over intra-chunk temporal dependencies.
% while the Top-$K$ operation is applied only to past frames for efficient history retrieval.

Given the selected frame set $\mathcal{T}_r$, we further perform fine-grained blockwise selection among all blocks in the selected frames.
For a frame $\tau \in \mathcal{T}_r$, let $\tilde{\vk}^{(\tau)}_j \in \mathbb{R}^{d}$ denote the $j$-th blockwise key summary. We compute block-level relevance logits as
\begin{equation}
o^{(i)}_r(\tau,j)
=
\langle \tilde{\vq}^{(i)}_r,\; \tilde{\vk}^{(\tau)}_j \rangle,
\label{eq:block_level_score}
\end{equation}
and select the top-$k$ block pairs from the candidate set $\mathcal{B}_r$:
\begin{equation}
\begin{aligned}
\mathcal{B}_r
&=
\{(\tau,j)\mid \tau\in\mathcal{T}_r,\; j\in\{1,\ldots,\lceil n/b_{kv}\rceil\}\},\\
\mathcal{J}_r
&=
\mathrm{TopK_{idx}}\Big(\{o^{(i)}_r(\tau,j)\}_{(\tau,j)\in\mathcal{B}_r}\Big).
\end{aligned}
\label{eq:block_selection}
\end{equation}

\noindent\textbf{Blockwise Sparse Attention.} Based on the selected frames and blocks, we construct a block-level attention mask $\mB^{(i)} \in \{0,1\}^{n_q \times n_{kv}}$. For the $r$-th query block, we have
\begin{equation}
\mB^{(i)}_r(\tau,j)
=
% \mathbf{1}\big[(\tau,j)\in\Omega_i\big].
\mathbf{1}\big[\tau\in\mathcal{T}_r,\; j\in\mathcal{J}_r(\tau)\big].
\label{eq:block_mask}
\end{equation}
Here, $n_q$ and $n_{kv}$ denote the number of query and key blocks, respectively. The final attention for the $r$-th query block is computed using blockwise sparse attention:
\begin{equation}
\mathrm{Attn}^{(i)}_r
=
\mathrm{softmax}\!\left(
\frac{\vq^{(i)}_r (\vk^{:i})^\top}{\sqrt{d}}
\odot \mM(\mB^{(i)}_r)
\right)\vv^{:i}.
\label{eq:block_sparse_attention}
\end{equation}

In summary, our HSA maintains a fixed attention complexity (independent of the total number of historical frames) while alleviating long-range consistency degradation. Meanwhile, our dual-stage mask selection incurs only a negligible overhead compared to conventional dynamic sparse attention, as it merely adds a frame-retrieval step (approximately a 2\% increase in end-to-end runtime). CAG and HSA are complementary: CAG allocates a sparsity ratio for each chunk at a macro level, while HSA determines from a fine-grained perspective how much historical information each block in the current chunk can leverage.

\begin{table*}[ht!]\setlength{\tabcolsep}{1pt}
 \renewcommand{\arraystretch}{1.05}
  \centering

  % \caption{Performance comparison with relevant baselines on $8$ dimensions of VBench~\cite{huang2024vbench}. ``+DMD2'' denotes our 4-step distilled \textsc{LinVideo} model. We highlight the best score and the second score in \textbf{bold} and \underline{underlined} formats, respectively.} 
   \caption{Performance comparison with state-of-the-art baselines on VBench~\cite{huang2024vbench}.} 
   % . Full results are provided in the Appendix.
  \resizebox{1\linewidth}{!}{
  \begin{tabular}[t!]{l|cc|cccccc|ccc}
\toprule
\multirow{2}{*}{Method}  & \multirow{2}{*}{Latency~(s)$\downarrow$} & \multirow{2}{*}{Speedup$\uparrow$} & \multirow{2}{*}{\rotatebox{0}{\makecell{Aesthetic\\Quality}$\uparrow$}} &  \multirow{2}{*}{\rotatebox{0}{\makecell{Imaging\\Quality}$\uparrow$}} & \multirow{2}{*}{\rotatebox{0}{\makecell{Motion\\Smoothness}$\uparrow$}} & \multirow{2}{*}{\rotatebox{0}{\makecell{Dynamic \\Degree}$\uparrow$}} & \multirow{2}{*}{\rotatebox{0}{\makecell{Subject\\Consistency}$\uparrow$}}  & \multirow{2}{*}{\rotatebox{0}{\makecell{Background \\Consistency}$\uparrow$}} & \multirow{2}{*}{\rotatebox{0}{\makecell{Quality\\Score}$\uparrow$}} & \multirow{2}{*}{\rotatebox{0}{\makecell{Semantic\\Score}$\uparrow$}} & \multirow{2}{*}{\rotatebox{0}{\makecell{Total\\Score}$\uparrow$}}\\
& & & & & & & & & & & \\
\midrule
\multicolumn{12}{c}{\cellcolor[gray]{0.92}Self-Forcing 1.3B ($\texttt{fps}=16$)} \\
\midrule
% 9612.0&1.00&67.4&70.0&98.3&63.1&95.3&96.5&84.8&81.2&84.1
% 8271.8&1.16&64.5&71.7&98.5&48.9&96.3&96.9&84.0&82.1&83.6
% 7399.3&1.30&45.8&66.1&96.0&88.6&90.2&93.6&78.7&53.7&73.7
% 21388.5&0.45&66.0&68.2&97.8&72.8&93.6&95.6&83.9&78.5&82.8
% 7710.0&1.25&66.7&69.8&98.3&44.2&95.6&96.7&83.4&82.5&83.2
% 7426.0&1.29&65.2&69.9&97.3&84.2&92.8&95.5&84.5&80.3&83.6
% 7394.6&1.30&67.2&71.0&98.3&66.7&96.2&96.5&85.4&80.9&84.5
 FlashAttention2~\cite{dao2023flashattention}  & 9.61  & $1.00\times$ &67.4&70.0&98.3&63.1&95.3&96.5&84.8&81.2&84.1  \\
 \midrule
 STA~\cite{zhang2025fast} & 8.27 & $1.16\times$ &  64.5&71.7&98.5&48.9&96.3&96.9&84.0&82.1&83.6\\
 Radial~\cite{li2025radial} &  7.39 & $1.30\times$ &45.8&66.1&96.0&88.6&90.2&93.6&78.7&53.7&73.7 \\
 SVG2~\cite{yang2025sparse} &  21.38 & $0.45\times$ & 66.0&68.2&97.8&72.8&93.6&95.6&83.9&78.5&82.8\\
 VMoBA~\cite{wu2025vmoba} &  7.42 & $1.29\times$ &65.2&69.9&97.3&84.2&92.8&95.5&84.5&80.3&83.6\\
 SLA~\cite{zhang2025sla} &  7.71 & $1.25\times$ & 66.7&69.8&98.3&44.2&95.6&96.7&83.4&82.5&83.2\\
\midrule
\rowcolor{mycolor!30}\textsc{Light Forcing} &  \textbf{7.39} & $\mathbf{1.30}\times$ & 67.2&71.0&98.3&66.7&96.2&96.5&85.4&80.9&\textbf{84.5} \\

\midrule
\multicolumn{12}{c}{\cellcolor[gray]{0.92}LongLive 1.3B ($\texttt{fps}=16$)} \\
\midrule
 FlashAttention2~\cite{dao2023flashattention}& 10.47 & $1.00\times$ & 68.7&69.3&98.8&39.2&97.0&97.2&83.8&80.7&83.2  \\
 \midrule
%  10472.5&1.00&97.0&97.2&68.7&69.3&98.8&39.2&83.8&80.7&83.2
% 9561.9&1.10&97.4&97.8&65.6&71.2&99.0&22.8&82.8&81.6&82.6
% 8894.3&1.18&77.6&88.9&55.1&72.0&98.0&25.0&75.0&66.6&73.3
% 22128.0&0.47&95.3&96.1&66.7&67.0&98.5&44.4&82.7&78.7&81.9
% 8818.7&1.19&58.3&80.9&59.9&68.2&97.5&50.6&71.5&70.7&71.3
% 8883.9&1.18&96.9&96.7&67.2&70.6&98.2&59.4&84.8&80.2&83.9

% 68.7&69.3&98.8&39.2&97.0&97.2&83.8&80.7&83.2
% 65.6&71.2&99.0&22.8&97.4&97.8&82.8&81.6&82.6
% 55.1&72.0&98.0&25.0&77.6&88.9&75.0&66.6&73.3
% 66.7&67.0&98.5&44.4&95.3&96.1&82.7&78.7&81.9
% 59.9&68.2&97.5&50.6&58.3&80.9&71.5&70.7&71.3
% 67.2&70.6&98.2&59.4&96.9&96.7&84.8&80.2&83.9
 STA~\cite{zhang2025fast} & 9.56 & $1.10\times$ & 65.6&71.2&99.0&22.8&97.4&97.8&82.8&81.6&82.6 \\
 Radial~\cite{li2025radial} & 8.89 & $1.18\times$ &55.1&72.0&98.0&25.0&77.6&88.9&75.0&66.6&73.3 \\
 SVG2~\cite{yang2025sparse} & 22.12 & $0.47\times$ & 66.7&67.0&98.5&44.4&95.3&96.1&82.7&78.7&81.9\\
 VMoBA~\cite{wu2025vmoba} & 8.88 & $1.18\times$ &  59.9&68.2&97.5&50.6&58.3&80.9&71.5&70.7&71.3 \\
\midrule
\rowcolor{mycolor!30}\textsc{Light Forcing} & \textbf{8.81} &$ \mathbf{1.19}\times$ &67.2&70.6&98.2&59.4&96.9&96.7&84.8&80.2& \textbf{83.9} \\
\bottomrule
\end{tabular}
}
    \label{tab:compare}
\end{table*}

\section{Experiments}
\subsection{Experimental Details}

\noindent\textbf{Implementation.} 
We build sparse attention on top of the currently open-sourced autoregressive video generation models, Self Forcing~\cite{huang2025self}, LongLive~\cite{yang2025longlive} and Infinite-Forcing~\cite{infinite-forcing}. Following them, we use a chunk size of three latent frames. For Finetunable methods (\ie, VMoBA~\cite{wu2025vmoba}, SLA~\cite{zhang2025sla} and \textsc{Light Forcing}), we perform extra post-training for 2,000 iterations based on their pre-trained weights. For latency evaluation, we adopt the SpargeAttention kernel~\cite{zhang2025spargeattention} for all methods (except Sparse VideoGen2~\cite{yang2025sparse} due to its variable block lengths, for which we use FlashInfer~\cite{ye2025flashinfer} as the inference backend), and report the measured latency on an RTX 5090 GPU. 

% The hyperparameters $s_{target}$ and $s_{base}$ are set to 0.9 and 0.98, respectively.

% on 8$\times$H100 GPUs

\noindent\textbf{Evaluation.} We use VBench~\cite{huang2024vbench} and VBench-Long~\cite{huang2025vbench++} to evaluate generation quality on 5/15-second videos across 16 dimensions. These dimensions include Subject Consistency, Background Consistency, Aesthetic Quality, Imaging Quality, Object Class, Multiple Objects, Color, Spatial Relationship, Scene, Temporal Style, Overall Consistency, Human Action, Temporal Flickering, Motion Smoothness, Dynamic Degree, and Appearance Style. We report a representative subset of these metrics in the main paper, and the complete results are reported in the appendix. Notably, we adopt the test prompts rewritten by Self Forcing~\cite{huang2025self}. For fair comparisons, we set the block size of all sparse attention methods to 64. We also adjust the resolution from $480\times 832$ to $512\times 768$, which avoids excessive padding overhead and potential non-equivalence introduced by certain methods (\eg, VMoBA~\cite{wu2025vmoba}).

\noindent\textbf{Baselines.} We compare \textsc{Light Forcing} with state-of-the-art sparse attention methods for bidirectional video generation models, covering static mask selection approaches (STA~\cite{zhang2025fast}, Sparse VideoGen2~\cite{yang2025sparse}, and Radial Attention~\cite{li2025radial}) as well as dynamic mask selection approaches (VMoBA~\cite{wu2025vmoba} and SLA~\cite{zhang2025sla}). To ensure a fair comparison, we set the sparsity ratio of all static methods to around 80\% (except for STA), and that of all dynamic methods to around 90\%. Additional implementation details and hyperparameter settings for the specific method are provided in the Appendix.
\begin{figure*}[ht!]
   \centering
        \includegraphics[width=0.9\textwidth]{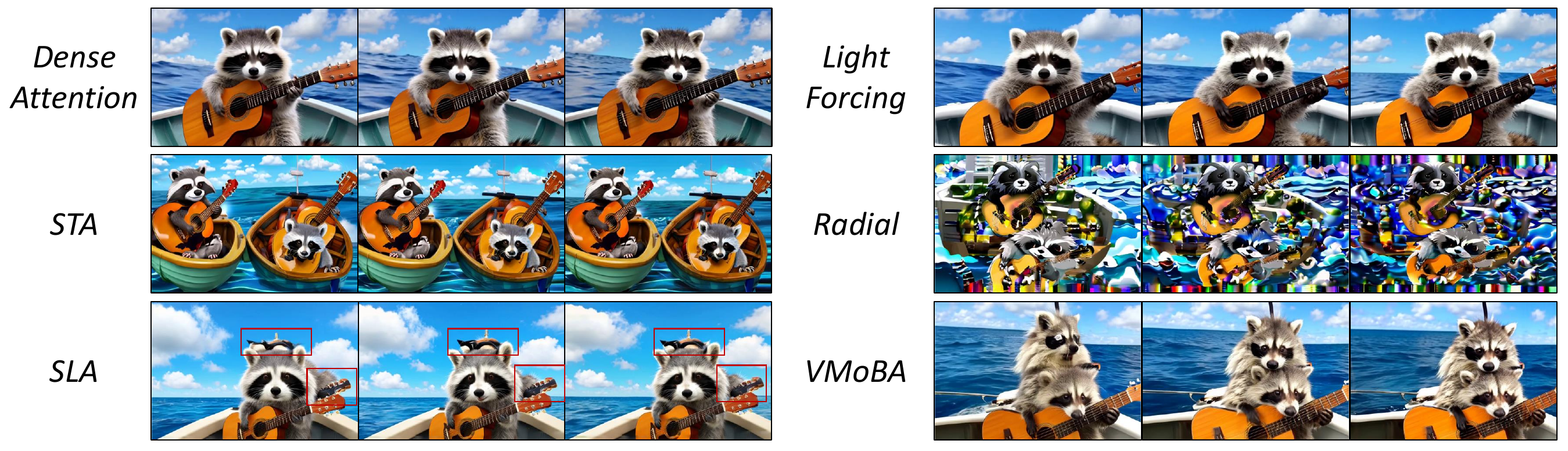}
        % \vspace{-0.1in}
        \caption{Qualitative comparisons of 5-second videos generated under the prompt ``A cute raccoon playing guitar in a boat on the ocean'' on Self Forcing~\cite{huang2025self}. We select frames at 0s, 2s, and 5s as representative snapshots of the video.}
        \label{fig:visual}
        % \vspace{-0.1in}
\end{figure*}
\subsection{Main Results}

Tab.~\ref{tab:compare} reports the evaluation results of our \textsc{Light Forcing} and state-of-the-art baselines on two mainstream autoregressive video generation models, \ie, Self Forcing~\cite{huang2025self} and LongLive~\cite{yang2025longlive}. We observe that \textsc{Light Forcing} outperforms these methods by a large margin on most metrics (\eg, Imaging Quality and Subject Consistency), while also achieving the highest speedups (1.3$\times$ on Self Forcing and 1.19$\times$ on LongLive). Notably, \textsc{Light Forcing} yields a higher Total Score than dense FlashAttention~\cite{dao2023flashattention} baselines (84.5 \emph{vs.} 84.1 for Self Forcing and 83.9 \emph{vs.} 83.2 for LongLive), suggesting substantial redundancy in dense attention and indicating that properly designed sparse solutions can achieve lossless performance.

Moreover, \textsc{Light Forcing} demonstrates strong versatility and applicability, primarily in three aspects compared to prior methods. \ding{192} \textit{Sparsity ratio.} Even with the smallest window size (\ie, $(3,3,3)$), STA~\cite{zhang2025fast} attains only 62.5\% sparsity under such output resolution, and thus yields limited speedup. \ding{193} \textit{Extra overhead.} While permutation-based methods such as SVG2~\cite{yang2025sparse} achieve relatively strong performance among training-free methods, they require repeated clustering initialization as the KV cache evolves, incurring particularly large extra overhead for few-step generators. \ding{194} \textit{Training difficulty.} LongLive~\cite{yang2025longlive} adopts LoRA~\cite{hu2022lora} to finetune, making it challenging for some chunk-agnostic finetunable methods (\eg, VMoBA~\cite{wu2025vmoba}) to converge. Even after fine-tuning, their performance still remains far from satisfactory.

Qualitative comparisons are shown in Fig.~\ref{fig:visual}. This further highlights that \textsc{Light Forcing} preserves high-fidelity and consistent video examples, whereas other baselines exhibit pronounced degradation, including object duplication in multi-object scenes (\eg, STA~\cite{zhang2025fast} and VMoBA~\cite{wu2025vmoba} producing two or more raccoons), anomalous objects (\eg, SLA~\cite{zhang2025sla} generating multiple handles of the guitar), and severe color shifts and artifacts (\eg, Radial~\cite{li2025radial}). These observations suggest that \textsc{Light Forcing} better mitigates error accumulation and over-exposure effects, enabling high-quality long-duration video synthesis.
% Additional video examples on LongLive~\cite{yang2025longlive} are in the Appendix.

\subsection{Longer Generation}

\begin{table*}[ht!]\setlength{\tabcolsep}{1pt}
 \renewcommand{\arraystretch}{1.05}
  \centering
  \caption{Long-video generation results on VBench-Long~\cite{huang2025vbench++}. We report representative metrics and the final scores for 15-second videos.}
  \resizebox{0.9\linewidth}{!}{
  \begin{tabular}[t!]{l|cccccc|ccc}
\toprule
\multirow{2}{*}{Method} & \multirow{2}{*}{\rotatebox{0}{\makecell{Aesthetic\\Quality}$\uparrow$}} &  \multirow{2}{*}{\rotatebox{0}{\makecell{Imaging\\Quality}$\uparrow$}} & \multirow{2}{*}{\rotatebox{0}{\makecell{Motion\\Smoothness}$\uparrow$}} & \multirow{2}{*}{\rotatebox{0}{\makecell{Dynamic\\Degree}$\uparrow$}} & \multirow{2}{*}{\rotatebox{0}{\makecell{Subject\\Consistency}$\uparrow$}} & \multirow{2}{*}{\rotatebox{0}{\makecell{Background\\Consistency}$\uparrow$}} & \multirow{2}{*}{\rotatebox{0}{\makecell{Quality\\Score}$\uparrow$}} & \multirow{2}{*}{\rotatebox{0}{\makecell{Semantic\\Score}$\uparrow$}} & \multirow{2}{*}{\rotatebox{0}{\makecell{Total\\Score}$\uparrow$}}\\
& & & & & & & & & \\
\midrule
\multicolumn{10}{c}{\cellcolor[gray]{0.92}Infinite-Forcing 1.3B  ($\texttt{fps}=16$)} \\
\midrule
 FlashAttention2~\cite{dao2023flashattention}  & 65.0 & 68.7 & 98.5 & 54.7 & 98.3 & 97.6 & 84.6 & 79.7 & 83.6 \\
\midrule
\rowcolor{mycolor!30}\textsc{Light Forcing} & 65.1 & 69.5 & 98.6 & 64.7 & 98.0 & 97.0 & 85.4 & 79.3 & 84.1 \\
\bottomrule
\end{tabular}
}
    \label{tab:long_generation}
\end{table*}

\begin{figure*}[ht!]
   \centering
        \includegraphics[width=1\textwidth]{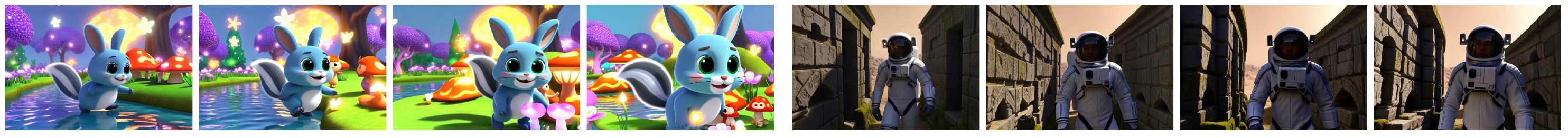}
        \vspace{-0.1in}
        \caption{Qualitative examples of 15-second videos generated using \textsc{Light Forcing}. Detailed prompts are provided in the Appendix.}
        \label{fig:long_gen_visual}
        \vspace{-0.1in}
\end{figure*}

For longer videos, since Self Forcing~\cite{huang2025self} does not natively support generation beyond 5 seconds, we evaluate \textsc{Light Forcing} with Infinite-Forcing~\cite{infinite-forcing} on 15-second videos using VBench-Long~\cite{huang2025vbench++}. As shown in Tab.~\ref{tab:long_generation}, \textsc{Light Forcing} still maintains a competitive and slightly higher Total Score than Infinite-Forcing (84.1 \emph{vs.} 83.6), suggesting that the proposed \textsc{Light Forcing} remains stable under longer autoregressive rollouts. In particular, \textsc{Light Forcing} improves Quality Score from 84.6 to 85.4, while achieving better Imaging Quality (69.5 \emph{vs.} 68.7), Motion Smoothness (98.6 \emph{vs.} 98.5), and Dynamic Degree (64.7 \emph{vs.} 54.7). The qualitative long-video results in Fig.~\ref{fig:long_gen_visual} further verify this observation, showing that \textsc{Light Forcing} preserves favorable motion smoothness and visual quality under 15-second generation. Complete VBench-Long results are provided in the Appendix.

% These gains indicate that reducing redundant attention computation does not amplify long-horizon error accumulation; instead, the chunk-aware sparsity schedule preserves visually important historical context while leaving sufficient capacity for dynamic evolution. Although Infinite-Forcing has slightly stronger Subject and Background Consistency, \textsc{Light Forcing} obtains comparable temporal stability and stronger motion dynamics, showing a favorable trade-off for efficient long-video generation.

\subsection{Ablation Studies}

\begin{table}[ht]\setlength{\tabcolsep}{1pt}
 \renewcommand{\arraystretch}{1.05}
  \centering
 
  \caption{ Ablation results for each component of \textsc{Light Forcing}. ``+1D Sparse Attention'' means directly applying sparse attention~\cite{zhang2025spargeattention} under the pretrained Self Forcing weights~\cite{huang2025self}.} 
  \resizebox{0.93\linewidth}{!}{
  
\begin{tabular}[t!]{l|cccc|c}
\toprule
\multirow{2}{*}{Method} 
& \multirow{2}{*}{\rotatebox{0}{\makecell{Subject\\Consistency}$\uparrow$}} 
& \multirow{2}{*}{\rotatebox{0}{\makecell{Aesthetic\\Quality}$\uparrow$}}  
& \multirow{2}{*}{\rotatebox{0}{\makecell{Imaging\\Quality}$\uparrow$}} 
& \multirow{2}{*}{\rotatebox{0}{\makecell{Dynamic\\Degree}$\uparrow$}} 
& \multirow{2}{*}{\rotatebox{0}{\makecell{Total\\Score}$\uparrow$}}\\
 & & & & & \\
\midrule

\rowcolor{mycolor!30}\textsc{Flash Attention} 
& 95.3 & 67.4 & 70.0 & 63.1 & 84.1 \\

\midrule
\textit{+1D Sparse Attention} 
& 86.9 & 51.4 & 66.0 & 52.8 & 73.0 \\

\textit{+Finetune} 
& 94.9 & 65.1 & 69.8 & 46.4 & 82.8 \\

\midrule
+ CAG
& 96.1 & 67.7 & 71.0 & 37.5 & 83.2 \\

+ CAG \& HSA 
& 96.2 & 67.2 & 71.0 & 66.7 & \textbf{84.5} \\

\bottomrule
\end{tabular}

}
    \label{tab:ablation}
\end{table}
\noindent \textbf{Ablation for Components.} We evaluate the effect of our two components on Self Forcing~\cite{huang2025self}. We observe that directly applying 1D sparse attention (90\% sparsity) without fine-tuning results in a severe quality collapse (\eg, degraded visual fidelity and dynamics). Although further fine-tuning partially recovers performance, it still falls short of dense attention (84.1 \emph{vs.} 82.8 in Total Score). When combined with CAG, the model exhibits notable gains in Aesthetic Quality and Imaging Quality, but its dynamics deteriorate, suggesting that under aggressive sparsity the model relies more heavily on priors from preceding chunks at the expense of motion. In contrast, introducing HSA substantially improves dynamics and ultimately surpasses dense attention in Total Score. To more intuitively demonstrate the role of each component, we also include long-video visualizations of CAG and HSA in the Appendix.

\begin{table}[ht]\setlength{\tabcolsep}{1pt}
  \centering
  \setlength{\tabcolsep}{4mm}

  \caption{Ablation study on the number of retrieved frames ($topk$) in Hierarchical Sparse Attention (HSA).} 
  % \vspace{-0.1in}
  \resizebox{0.8\linewidth}{!}{
  \begin{tabular}[t!]{c|cc|c}
\toprule
\multirow{2}{*}{\texttt{Top-k}} 
& \multirow{2}{*}{\makecell{Quality\\Score}$\uparrow$} 
& \multirow{2}{*}{\makecell{Semantic\\Score}$\uparrow$} 
& \multirow{2}{*}{\makecell{Total\\Score}$\uparrow$} \\
& & & \\
\midrule
6  & \textbf{85.4} & \textbf{80.9} & \textbf{84.5} \\
9  & 85.2 & 80.8 & 84.4 \\
12 & 85.1 & \textbf{80.9} & 84.3 \\
\bottomrule
\end{tabular}
}
    \label{tab:ablation_hsa}
    % \vspace{-0.1in}
\end{table}

\noindent \textbf{Sensitivity Analysis on HSA.} We conduct a hyperparameter study on the number of retrieved frames ($topk$) in the first-stage retrieval of HSA, reported in Tab.~\ref{tab:ablation_hsa}. Due to limited time and resources, we evaluate three settings ($topk\in\{6,9,12\}$). Our method remains highly robust, achieving similarly strong performance across these choices. This further suggests that, for each query block, attending to only a small subset of past frames is sufficient to mitigate inconsistency issues.

% In contrast, combining either Method~1 or Method~2 significantly improves consistency and dynamics, and finally surpasses dense attention in Total Score.

\begin{figure}[t!]
   \centering
   % \vspace{-0.1in}
    \setlength{\abovecaptionskip}{0.2cm}
        \includegraphics[width=0.48\textwidth]{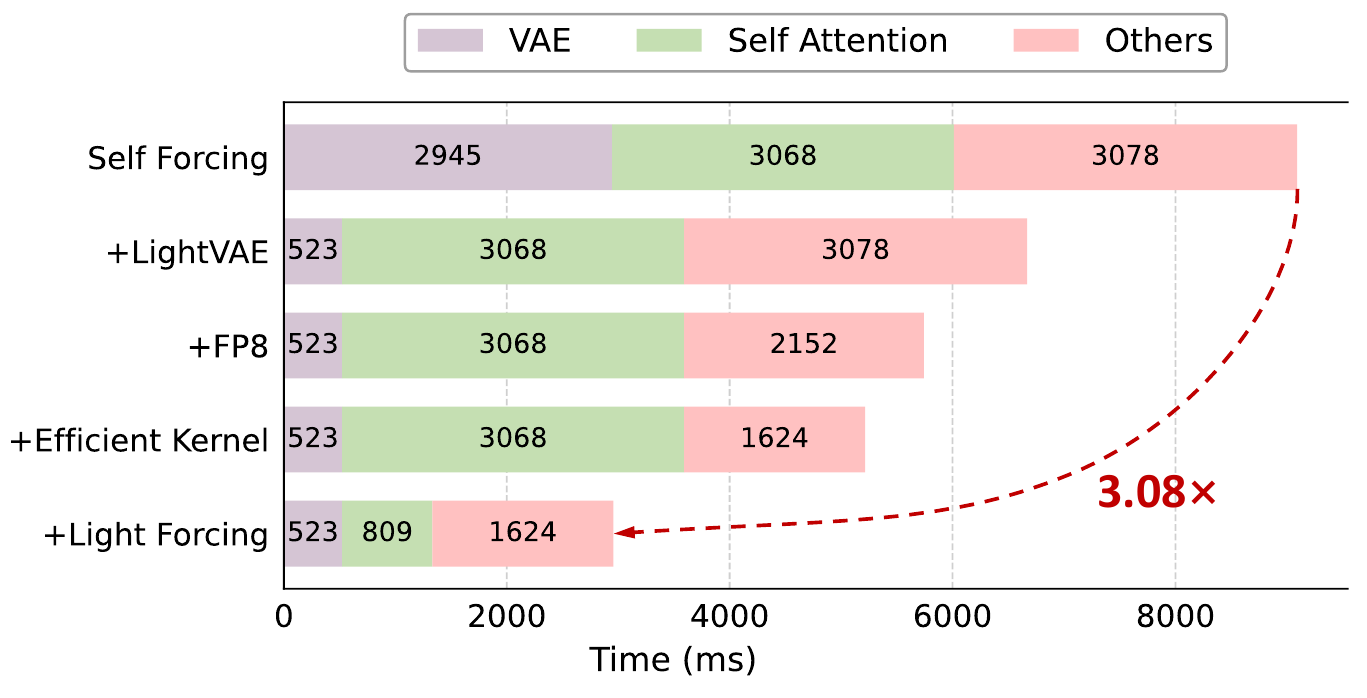}
        % \vspace{-0.2in}
        \caption{Efficient deployment of Light Forcing. We measure its latency on RTX~5090 for 5-second video generation.}
        \label{fig:time}
        % \vspace{-0.2in}
\end{figure}
\subsection{Efficient Deployment}

% To better unlock the acceleration potential of autoregressive video generation models, we deploy \textsc{Light Forcing} on the mainstream video generation inference framework LightX2V~\cite{lightx2v} and profile the runtime latency of each component (see in Fig.~\ref{fig:time}). We replace the default Wan VAE~\cite{wan2025wan} with efficient LightVAE~\cite{lightx2v}, and further deploy all linear layers in the model using low-bit FP8 precision (We quantize weights with per-channel granularity and activations with per-token granularity). Both choices are widely regarded as lossless acceleration techniques. 

% In our latency evaluation, \textsc{Light Forcing} achieves a $3.29\times$ speedup in attention time and a $2.33\times$ end-to-end speedup, while maintaining satisfactory generation quality. Remarkably, \textsc{Light Forcing} 1.3B achieves 19.7\,FPS, enabling real-time generation on a consumer-level GPU for the first time.
To better unlock the acceleration potential of autoregressive video generation models, we deploy \textsc{Light Forcing} on the mainstream video generation inference framework LightX2V~\cite{lightx2v} and profile the runtime latency of each component (see Fig.~\ref{fig:time}). We replace the default Wan VAE~\cite{wan2025wan} with the efficient LightVAE~\cite{lightx2v}, and further deploy all linear layers in the model using low-bit FP8 precision, where weights are quantized with per-channel granularity and activations with per-token granularity. Both choices are widely adopted as nearly lossless acceleration techniques. In addition, we provide several Triton-based optimized kernels, including RoPE, RMSNorm, and \texttt{fuse\_scale\_shift}, to further reduce operator overhead in the end-to-end inference pipeline. In our latency evaluation, measured with our open-source Docker image\footnote{\texttt{lvchengtao/light\_forcing:v1}.}, \textsc{Light Forcing} achieves a $3.79\times$ speedup in attention time and a $3.08\times$ end-to-end speedup, while maintaining satisfactory generation quality. Remarkably, \textsc{Light Forcing} 1.3B achieves 27.4\,FPS, enabling real-time video generation on a consumer-grade GPU for the first time. Additional latency evaluations on other GPU platforms (\eg, H100 and A100) and longer durations are provided in the Appendix.

\section{Conclusion}

We proposed \textsc{Light Forcing}, a sparse attention framework tailored for autoregressive video diffusion. By introducing chunk-aware growth and hierarchical sparse attention, our method effectively mitigates error accumulation while preserving long-range context. Extensive experiments demonstrate consistent improvements in both efficiency and generation quality, enabling real-time video synthesis on consumer GPUs and establishing a strong foundation for scalable AR video generation. 

% Despite the strong empirical results, our work has several limitations. First, evaluations are conducted on 1.3B models. Scaling \textsc{Light Forcing} to larger models, such as realtime-video 14B model, remains an important direction for future study. Second, our sparse attention implementation is based on FlashAttention 2~\cite{dao2023flashattention}, which requires additional kernel adaptations for newer GPU architectures (\eg, Hopper). Finally, while sparsity proves highly effective, how to further combine it with other acceleration techniques (\eg, step distillation or low-bit quantization) to unlock greater performance gains remains an open and promising research direction.

% In the unusual situation where you want a paper to appear in the
% references without citing it in the main text, use \nocite
% \nocite{langley00}

\section*{Acknowledgements}
This research/project is supported by A*STAR under the RIE2025 Industry Alignment Fund – Industry Collaboration Projects (IAF-ICP) Funding Initiative (Award: I2501E0045), as well as cash and in-kind contribution from the industry partner(s). This research/project is also supported by the NTU Start-Up Grant, Singapore.

\section*{Impact Statement}
This paper presents work whose goal is to advance the field of Machine Learning. There are many potential societal consequences of our work, none of which we feel must be specifically highlighted here.
% This paper proposes an efficient autoregressive video generation model that significantly reduces latency through a redesigned attention mechanism while maintaining generation quality. We will open-source the related code after the paper is published to promote transparency, reproducibility, and further research in efficient artificial intelligence. We also acknowledge that video generation may, in principle, be misused. Therefore, we encourage the community to consider and develop guidelines for the ethical use of such generative technologies.

\bibliography{example_paper}

@article{xu2025xattention,
  title={Xattention: Block sparse attention with antidiagonal scoring},
  author={Xu, Ruyi and Xiao, Guangxuan and Huang, Haofeng and Guo, Junxian and Han, Song},
  journal={arXiv preprint arXiv:2503.16428},
  year={2025}
}

@inproceedings{zhang2025spargeattention,
  title={SpargeAttention: Accurate and Training-free Sparse Attention Accelerating Any Model Inference},
  author={Zhang, Jintao and Xiang, Chendong and Huang, Haofeng and Wei, Jia and Xi, Haocheng and Zhu, Jun and Chen, Jianfei},
  booktitle={International Conference on Machine Learning},
  pages={76397--76413},
  year={2025},
  organization={PMLR}
}

@article{zhang2025sla,
  title={SLA: Beyond Sparsity in Diffusion Transformers via Fine-Tunable Sparse-Linear Attention},
  author={Zhang, Jintao and Wang, Haoxu and Jiang, Kai and Yang, Shuo and Zheng, Kaiwen and Xi, Haocheng and Wang, Ziteng and Zhu, Hongzhou and Zhao, Min and Stoica, Ion and others},
  journal={arXiv preprint arXiv:2509.24006},
  year={2025}
}

@article{huang2025linvideo,
  title={Linvideo: A post-training framework towards o (n) attention in efficient video generation},
  author={Huang, Yushi and Ge, Xingtong and Gong, Ruihao and Lv, Chengtao and Zhang, Jun},
  journal={arXiv preprint arXiv:2510.08318},
  year={2025}
}

@article{xi2025sparse,
  title={Sparse videogen: Accelerating video diffusion transformers with spatial-temporal sparsity},
  author={Xi, Haocheng and Yang, Shuo and Zhao, Yilong and Xu, Chenfeng and Li, Muyang and Li, Xiuyu and Lin, Yujun and Cai, Han and Zhang, Jintao and Li, Dacheng and others},
  journal={arXiv preprint arXiv:2502.01776},
  year={2025}
}

@article{yang2025sparse,
  title={Sparse VideoGen2: Accelerate Video Generation with Sparse Attention via Semantic-Aware Permutation},
  author={Yang, Shuo and Xi, Haocheng and Zhao, Yilong and Li, Muyang and Zhang, Jintao and Cai, Han and Lin, Yujun and Li, Xiuyu and Xu, Chenfeng and Peng, Kelly and others},
  journal={arXiv preprint arXiv:2505.18875},
  year={2025}
}

@article{wu2025vmoba,
  title={VMoBA: Mixture-of-Block Attention for Video Diffusion Models},
  author={Wu, Jianzong and Hou, Liang and Yang, Haotian and Tao, Xin and Tian, Ye and Wan, Pengfei and Zhang, Di and Tong, Yunhai},
  journal={arXiv preprint arXiv:2506.23858},
  year={2025}
}

@article{zhang2025fast,
  title={Fast video generation with sliding tile attention},
  author={Zhang, Peiyuan and Chen, Yongqi and Su, Runlong and Ding, Hangliang and Stoica, Ion and Liu, Zhengzhong and Zhang, Hao},
  journal={arXiv preprint arXiv:2502.04507},
  year={2025}
}

@article{zhang2025vsa,
  title={Vsa: Faster video diffusion with trainable sparse attention},
  author={Zhang, Peiyuan and Chen, Yongqi and Huang, Haofeng and Lin, Will and Liu, Zhengzhong and Stoica, Ion and Xing, Eric and Zhang, Hao},
  journal={arXiv preprint arXiv:2505.13389},
  year={2025}
}

@article{li2025radial,
  title={Radial Attention: $O(n\log n)$ Sparse Attention with Energy Decay for Long Video Generation},
  author={Li, Xingyang and Li, Muyang and Cai, Tianle and Xi, Haocheng and Yang, Shuo and Lin, Yujun and Zhang, Lvmin and Yang, Songlin and Hu, Jinbo and Peng, Kelly and others},
  journal={arXiv preprint arXiv:2506.19852},
  year={2025}
}

@article{hassani2025generalized,
  title={Generalized Neighborhood Attention: Multi-dimensional Sparse Attention at the Speed of Light},
  author={Hassani, Ali and Zhou, Fengzhe and Kane, Aditya and Huang, Jiannan and Chen, Chieh-Yun and Shi, Min and Walton, Steven and Hoehnerbach, Markus and Thakkar, Vijay and Isaev, Michael and others},
  journal={arXiv preprint arXiv:2504.16922},
  year={2025}
}

@article{li2025psa,
  title={PSA: Pyramid Sparse Attention for Efficient Video Understanding and Generation},
  author={Li, Xiaolong and Gu, Youping and Lin, Xi and Wang, Weijie and Zhuang, Bohan},
  journal={arXiv preprint arXiv:2512.04025},
  year={2025}
}

@article{zhou2025trainable,
  title={Trainable Log-linear Sparse Attention for Efficient Diffusion Transformers},
  author={Zhou, Yifan and Xiao, Zeqi and Wei, Tianyi and Yang, Shuai and Pan, Xingang},
  journal={arXiv preprint arXiv:2512.16615},
  year={2025}
}

@article{wan2025wan,
  title={Wan: Open and advanced large-scale video generative models},
  author={Wan, Team and Wang, Ang and Ai, Baole and Wen, Bin and Mao, Chaojie and Xie, Chen-Wei and Chen, Di and Yu, Feiwu and Zhao, Haiming and Yang, Jianxiao and others},
  journal={arXiv preprint arXiv:2503.20314},
  year={2025}
}

@article{sun2024hunyuan,
  title={Hunyuan-large: An open-source moe model with 52 billion activated parameters by tencent},
  author={Sun, Xingwu and Chen, Yanfeng and Huang, Yiqing and Xie, Ruobing and Zhu, Jiaqi and Zhang, Kai and Li, Shuaipeng and Yang, Zhen and Han, Jonny and Shu, Xiaobo and others},
  journal={arXiv preprint arXiv:2411.02265},
  year={2024}
}

@article{yang2024cogvideox,
  title={Cogvideox: Text-to-video diffusion models with an expert transformer},
  author={Yang, Zhuoyi and Teng, Jiayan and Zheng, Wendi and Ding, Ming and Huang, Shiyu and Xu, Jiazheng and Yang, Yuanming and Hong, Wenyi and Zhang, Xiaohan and Feng, Guanyu and others},
  journal={arXiv preprint arXiv:2408.06072},
  year={2024}
}

@article{chen2025skyreels,
  title={Skyreels-v2: Infinite-length film generative model},
  author={Chen, Guibin and Lin, Dixuan and Yang, Jiangping and Lin, Chunze and Zhu, Junchen and Fan, Mingyuan and Zhang, Hao and Chen, Sheng and Chen, Zheng and Ma, Chengcheng and others},
  journal={arXiv preprint arXiv:2504.13074},
  year={2025}
}

@article{teng2025magi,
  title={MAGI-1: Autoregressive Video Generation at Scale},
  author={Teng, Hansi and Jia, Hongyu and Sun, Lei and Li, Lingzhi and Li, Maolin and Tang, Mingqiu and Han, Shuai and Zhang, Tianning and Zhang, WQ and Luo, Weifeng and others},
  journal={arXiv preprint arXiv:2505.13211},
  year={2025}
}

@article{gao2024ca2,
  title={Ca2-vdm: Efficient autoregressive video diffusion model with causal generation and cache sharing},
  author={Gao, Kaifeng and Shi, Jiaxin and Zhang, Hanwang and Wang, Chunping and Xiao, Jun and Chen, Long},
  journal={arXiv preprint arXiv:2411.16375},
  year={2024}
}

@article{hu2024acdit,
  title={Acdit: Interpolating autoregressive conditional modeling and diffusion transformer},
  author={Hu, Jinyi and Hu, Shengding and Song, Yuxuan and Huang, Yufei and Wang, Mingxuan and Zhou, Hao and Liu, Zhiyuan and Ma, Wei-Ying and Sun, Maosong},
  journal={arXiv preprint arXiv:2412.07720},
  year={2024}
}

@article{liu2025rolling,
  title={Rolling forcing: Autoregressive long video diffusion in real time},
  author={Liu, Kunhao and Hu, Wenbo and Xu, Jiale and Shan, Ying and Lu, Shijian},
  journal={arXiv preprint arXiv:2509.25161},
  year={2025}
}

@article{cui2025self,
  title={Self-Forcing++: Towards Minute-Scale High-Quality Video Generation},
  author={Cui, Justin and Wu, Jie and Li, Ming and Yang, Tao and Li, Xiaojie and Wang, Rui and Bai, Andrew and Ban, Yuanhao and Hsieh, Cho-Jui},
  journal={arXiv preprint arXiv:2510.02283},
  year={2025}
}

@article{zhang2026frame,
  title={Frame context packing and drift prevention in next-frame-prediction video diffusion models},
  author={Zhang, Lvmin and Cai, Shengqu and Li, Muyang and Wetzstein, Gordon and Agrawala, Maneesh},
  journal={Advances in Neural Information Processing Systems},
  volume={38},
  pages={30546--30566},
  year={2026}
}

@article{yang2025longlive,
  title={Longlive: Real-time interactive long video generation},
  author={Yang, Shuai and Huang, Wei and Chu, Ruihang and Xiao, Yicheng and Zhao, Yuyang and Wang, Xianbang and Li, Muyang and Xie, Enze and Chen, Yingcong and Lu, Yao and others},
  journal={arXiv preprint arXiv:2509.22622},
  year={2025}
}

@article{lu2025reward,
  title={Reward Forcing: Efficient Streaming Video Generation with Rewarded Distribution Matching Distillation},
  author={Lu, Yunhong and Zeng, Yanhong and Li, Haobo and Ouyang, Hao and Wang, Qiuyu and Cheng, Ka Leong and Zhu, Jiapeng and Cao, Hengyuan and Zhang, Zhipeng and Zhu, Xing and others},
  journal={arXiv preprint arXiv:2512.04678},
  year={2025}
}

@inproceedings{yin2025slow,
  title={From slow bidirectional to fast autoregressive video diffusion models},
  author={Yin, Tianwei and Zhang, Qiang and Zhang, Richard and Freeman, William T and Durand, Fredo and Shechtman, Eli and Huang, Xun},
  booktitle={Proceedings of the Computer Vision and Pattern Recognition Conference},
  pages={22963--22974},
  year={2025}
}

@article{huang2025self,
  title={Self Forcing: Bridging the Train-Test Gap in Autoregressive Video Diffusion},
  author={Huang, Xun and Li, Zhengqi and He, Guande and Zhou, Mingyuan and Shechtman, Eli},
  journal={arXiv preprint arXiv:2506.08009},
  year={2025}
}

@article{chen2024diffusion,
  title={Diffusion forcing: Next-token prediction meets full-sequence diffusion},
  author={Chen, Boyuan and Mart{\'\i} Mons{\'o}, Diego and Du, Yilun and Simchowitz, Max and Tedrake, Russ and Sitzmann, Vincent},
  journal={Advances in Neural Information Processing Systems},
  volume={37},
  pages={24081--24125},
  year={2024}
}

@article{yin2024improved,
  title={Improved distribution matching distillation for fast image synthesis},
  author={Yin, Tianwei and Gharbi, Micha{\"e}l and Park, Taesung and Zhang, Richard and Shechtman, Eli and Durand, Fredo and Freeman, Bill},
  journal={Advances in neural information processing systems},
  volume={37},
  pages={47455--47487},
  year={2024}
}

@inproceedings{henschel2025streamingt2v,
  title={Streamingt2v: Consistent, dynamic, and extendable long video generation from text},
  author={Henschel, Roberto and Khachatryan, Levon and Poghosyan, Hayk and Hayrapetyan, Daniil and Tadevosyan, Vahram and Wang, Zhangyang and Navasardyan, Shant and Shi, Humphrey},
  booktitle={Proceedings of the Computer Vision and Pattern Recognition Conference},
  pages={2568--2577},
  year={2025}
}

@article{gu2025long,
  title={Long-context autoregressive video modeling with next-frame prediction},
  author={Gu, Yuchao and Mao, Weijia and Shou, Mike Zheng},
  journal={arXiv preprint arXiv:2503.19325},
  year={2025}
}

@article{kodaira2025streamdit,
  title={Streamdit: Real-time streaming text-to-video generation},
  author={Kodaira, Akio and Hou, Tingbo and Hou, Ji and Georgopoulos, Markos and Juefei-Xu, Felix and Tomizuka, Masayoshi and Zhao, Yue},
  journal={arXiv preprint arXiv:2507.03745},
  year={2025}
}

@article{zhang2024sageattention,
  title={Sageattention: Accurate 8-bit attention for plug-and-play inference acceleration},
  author={Zhang, Jintao and Wei, Jia and Huang, Haofeng and Zhang, Pengle and Zhu, Jun and Chen, Jianfei},
  journal={arXiv preprint arXiv:2410.02367},
  year={2024}
}

@article{zhang2024sageattention2,
  title={Sageattention2: Efficient attention with thorough outlier smoothing and per-thread int4 quantization},
  author={Zhang, Jintao and Huang, Haofeng and Zhang, Pengle and Wei, Jia and Zhu, Jun and Chen, Jianfei},
  journal={arXiv preprint arXiv:2411.10958},
  year={2024}
}

@article{chen2025sana,
  title={Sana-video: Efficient video generation with block linear diffusion transformer},
  author={Chen, Junsong and Zhao, Yuyang and Yu, Jincheng and Chu, Ruihang and Chen, Junyu and Yang, Shuai and Wang, Xianbang and Pan, Yicheng and Zhou, Daquan and Ling, Huan and others},
  journal={arXiv preprint arXiv:2509.24695},
  year={2025}
}

@article{xie2024sana,
  title={Sana: Efficient high-resolution image synthesis with linear diffusion transformers},
  author={Xie, Enze and Chen, Junsong and Chen, Junyu and Cai, Han and Tang, Haotian and Lin, Yujun and Zhang, Zhekai and Li, Muyang and Zhu, Ligeng and Lu, Yao and others},
  journal={arXiv preprint arXiv:2410.10629},
  year={2024}
}

@inproceedings{huang2024vbench,
  title={Vbench: Comprehensive benchmark suite for video generative models},
  author={Huang, Ziqi and He, Yinan and Yu, Jiashuo and Zhang, Fan and Si, Chenyang and Jiang, Yuming and Zhang, Yuanhan and Wu, Tianxing and Jin, Qingyang and Chanpaisit, Nattapol and others},
  booktitle={Proceedings of the IEEE/CVF Conference on Computer Vision and Pattern Recognition},
  pages={21807--21818},
  year={2024}
}

@article{ye2025flashinfer,
    title = {FlashInfer: Efficient and Customizable Attention Engine for LLM Inference Serving},
    author = {
      Ye, Zihao and
      Chen, Lequn and
      Lai, Ruihang and
      Lin, Wuwei and
      Zhang, Yineng and
      Wang, Stephanie and
      Chen, Tianqi and
      Kasikci, Baris and
      Grover, Vinod and
      Krishnamurthy, Arvind and
      Ceze, Luis
    },
    journal = {arXiv preprint arXiv:2501.01005},
    year = {2025},
    url = {https://arxiv.org/abs/2501.01005}
}

@inproceedings{peebles2023scalable,
  title={Scalable diffusion models with transformers},
  author={Peebles, William and Xie, Saining},
  booktitle={Proceedings of the IEEE/CVF international conference on computer vision},
  pages={4195--4205},
  year={2023}
}

@article{van2017neural,
  title={Neural discrete representation learning},
  author={Van Den Oord, Aaron and Vinyals, Oriol and others},
  journal={Advances in neural information processing systems},
  volume={30},
  year={2017}
}

@article{decart2024oasis,
  title={Oasis: A universe in a transformer},
  author={Decart, Etched and McIntyre, Quinn and Campbell, Spruce and Chen, Xinlei and Wachen, Robert},
  journal={URL: https://oasis-model. github. io},
  year={2024}
}

@article{parker2024genie,
  title={Genie 2: A large-scale foundation world model},
  author={Parker-Holder, J and Ball, P and Bruce, J and Dasagi, V and Holsheimer, K and Kaplanis, C and Moufarek, A and Scully, G and Shar, J and Shi, J and others},
  journal={URL: https://deepmind. google/discover/blog/genie-2-a-large-scale-foundation-world-model},
  year={2024}
}

@inproceedings{bruce2024genie,
  title={Genie: Generative interactive environments},
  author={Bruce, Jake and Dennis, Michael D and Edwards, Ashley and Parker-Holder, Jack and Shi, Yuge and Hughes, Edward and Lai, Matthew and Mavalankar, Aditi and Steigerwald, Richie and Apps, Chris and others},
  booktitle={Forty-first International Conference on Machine Learning},
  year={2024}
}

@article{li2025unified,
  title={Unified video action model},
  author={Li, Shuang and Gao, Yihuai and Sadigh, Dorsa and Song, Shuran},
  journal={arXiv preprint arXiv:2503.00200},
  year={2025}
}

@article{yang2023learning,
  title={Learning interactive real-world simulators},
  author={Yang, Mengjiao and Du, Yilun and Ghasemipour, Kamyar and Tompson, Jonathan and Schuurmans, Dale and Abbeel, Pieter},
  journal={arXiv preprint arXiv:2310.06114},
  volume={1},
  number={2},
  pages={6},
  year={2023}
}

@article{dao2023flashattention,
  title={Flashattention-2: Faster attention with better parallelism and work partitioning},
  author={Dao, Tri},
  journal={arXiv preprint arXiv:2307.08691},
  year={2023}
}

@inproceedings{liu2025timestep,
  title={Timestep Embedding Tells: It's Time to Cache for Video Diffusion Model},
  author={Liu, Feng and Zhang, Shiwei and Wang, Xiaofeng and Wei, Yujie and Qiu, Haonan and Zhao, Yuzhong and Zhang, Yingya and Ye, Qixiang and Wan, Fang},
  booktitle={Proceedings of the Computer Vision and Pattern Recognition Conference},
  pages={7353--7363},
  year={2025}
}

@article{ma2024learning,
  title={Learning-to-cache: Accelerating diffusion transformer via layer caching},
  author={Ma, Xinyin and Fang, Gongfan and Bi Mi, Michael and Wang, Xinchao},
  journal={Advances in Neural Information Processing Systems},
  volume={37},
  pages={133282--133304},
  year={2024}
}

@article{huang2024harmonica,
  title={HarmoniCa: Harmonizing Training and Inference for Better Feature Caching in Diffusion Transformer Acceleration},
  author={Huang, Yushi and Wang, Zining and Gong, Ruihao and Liu, Jing and Zhang, Xinjie and Guo, Jinyang and Liu, Xianglong and Zhang, Jun},
  journal={arXiv preprint arXiv:2410.01723},
  year={2024}
}

@article{hu2022lora,
  title={Lora: Low-rank adaptation of large language models.},
  author={Hu, Edward J and Shen, Yelong and Wallis, Phillip and Allen-Zhu, Zeyuan and Li, Yuanzhi and Wang, Shean and Wang, Lu and Chen, Weizhu and others},
  journal={ICLR},
  volume={1},
  number={2},
  pages={3},
  year={2022}
}

@article{li2023towards,
  title={Towards faster non-asymptotic convergence for diffusion-based generative models},
  author={Li, Gen and Wei, Yuting and Chen, Yuxin and Chi, Yuejie},
  journal={arXiv preprint arXiv:2306.09251},
  year={2023}
}

@article{beltagy2020longformer,
  title={Longformer: The long-document transformer},
  author={Beltagy, Iz and Peters, Matthew E and Cohan, Arman},
  journal={arXiv preprint arXiv:2004.05150},
  year={2020}
}

@misc{lightx2v,
 author = {LightX2V Contributors},
 title = {LightX2V: Light Video Generation Inference Framework},
 year = {2025},
 publisher = {GitHub},
 journal = {GitHub repository},
 howpublished = {\url{https://github.com/ModelTC/lightx2v}},
}

@software{krea_realtime_14b,
  title={Krea Realtime 14B: Real-time Video Generation},
  author={Erwann Millon},
  year={2025},
  url={https://github.com/krea-ai/realtime-video}
}

@inproceedings{song2023consistency,
  title={Consistency models},
  author={Song, Yang and Dhariwal, Prafulla and Chen, Mark and Sutskever, Ilya},
  booktitle={Proceedings of the 40th International Conference on Machine Learning},
  pages={32211--32252},
  year={2023}
}

@inproceedings{tao2025dycoke,
  title={Dycoke: Dynamic compression of tokens for fast video large language models},
  author={Tao, Keda and Qin, Can and You, Haoxuan and Sui, Yang and Wang, Huan},
  booktitle={Proceedings of the Computer Vision and Pattern Recognition Conference},
  pages={18992--19001},
  year={2025}
}

@inproceedings{lv2026llmc+,
  title={Llmc+: Benchmarking vision-language model compression with a plug-and-play toolkit},
  author={Lv, Chengtao and Zhang, Bilang and Yong, Yang and Gong, Ruihao and Huang, Yushi and Gu, Shiqiao and Wu, Jiajun and Shi, Yumeng and Guo, Jinyang and Wang, Wenya},
  booktitle={Proceedings of the AAAI Conference on Artificial Intelligence},
  volume={40},
  number={29},
  pages={24189--24197},
  year={2026}
}

@article{huang2025vbench++,
    title={{VBench++}: Comprehensive and Versatile Benchmark Suite for Video Generative Models},
    author={Huang, Ziqi and Zhang, Fan and Xu, Xiaojie and He, Yinan and Yu, Jiashuo and Dong, Ziyue and Ma, Qianli and Chanpaisit, Nattapol and Si, Chenyang and Jiang, Yuming and Wang, Yaohui and Chen, Xinyuan and Chen, Ying-Cong and Wang, Limin and Lin, Dahua and Qiao, Yu and Liu, Ziwei},
    journal={IEEE Transactions on Pattern Analysis and Machine Intelligence}, 
    year={2025},
    doi={10.1109/TPAMI.2025.3633890}
}

@misc{infinite-forcing,
    Author = {Junyi Chen, Zhoujie Fu, Xianglong He},
    Year = {2025},
    Note = {https://github.com/SOTAMak1r/Infinite-Forcing},
    Title = {Infinite-Forcing: Towards Infinite-Long Video Generation}
}

@article{cai2025mixture,
  title={Mixture of contexts for long video generation},
  author={Cai, Shengqu and Yang, Ceyuan and Zhang, Lvmin and Guo, Yuwei and Xiao, Junfei and Yang, Ziyan and Xu, Yinghao and Yang, Zhenheng and Yuille, Alan and Guibas, Leonidas and others},
  journal={arXiv preprint arXiv:2508.21058},
  year={2025}
}

@article{wu2025pack,
  title={Pack and force your memory: Long-form and consistent video generation},
  author={Wu, Xiaofei and Zhang, Guozhen and Xu, Zhiyong and Zhou, Yuan and Lu, Qinglin and He, Xuming},
  journal={arXiv preprint arXiv:2510.01784},
  year={2025}
}

@inproceedings{dalal2025one,
  title={One-minute video generation with test-time training},
  author={Dalal, Karan and Koceja, Daniel and Xu, Jiarui and Zhao, Yue and Han, Shihao and Cheung, Ka Chun and Kautz, Jan and Choi, Yejin and Sun, Yu and Wang, Xiaolong},
  booktitle={Proceedings of the Computer Vision and Pattern Recognition Conference},
  pages={17702--17711},
  year={2025}
}

@article{zadouri2026flashattention,
  title={Flashattention-4: Algorithm and kernel pipelining co-design for asymmetric hardware scaling},
  author={Zadouri, Ted and Hoehnerbach, Markus and Shah, Jay and Liu, Timmy and Thakkar, Vijay and Dao, Tri},
  journal={arXiv preprint arXiv:2603.05451},
  year={2026}
}

@inproceedings{yang2021r3det,
  title={R3det: Refined single-stage detector with feature refinement for rotating object},
  author={Yang, Xue and Yan, Junchi and Feng, Ziming and He, Tao},
  booktitle={Proceedings of the AAAI conference on artificial intelligence},
  volume={35},
  number={4},
  pages={3163--3171},
  year={2021}
}

@inproceedings{yang2019deep,
  title={Deep spectral clustering using dual autoencoder network},
  author={Yang, Xu and Deng, Cheng and Zheng, Feng and Yan, Junchi and Liu, Wei},
  booktitle={Proceedings of the IEEE/CVF conference on computer vision and pattern recognition},
  pages={4066--4075},
  year={2019}
}

@inproceedings{li2018self,
  title={Self-supervised adversarial hashing networks for cross-modal retrieval},
  author={Li, Chao and Deng, Cheng and Li, Ning and Liu, Wei and Gao, Xinbo and Tao, Dacheng},
  booktitle={Proceedings of the IEEE conference on computer vision and pattern recognition},
  pages={4242--4251},
  year={2018}
}

@article{deng2018triplet,
  title={Triplet-based deep hashing network for cross-modal retrieval},
  author={Deng, Cheng and Chen, Zhaojia and Liu, Xianglong and Gao, Xinbo and Tao, Dacheng},
  journal={IEEE Transactions on Image Processing},
  volume={27},
  number={8},
  pages={3893--3903},
  year={2018},
  publisher={IEEE}
}

@inproceedings{gong2019differentiable,
  title={Differentiable soft quantization: Bridging full-precision and low-bit neural networks},
  author={Gong, Ruihao and Liu, Xianglong and Jiang, Shenghu and Li, Tianxiang and Hu, Peng and Lin, Jiazhen and Yu, Fengwei and Yan, Junjie},
  booktitle={Proceedings of the IEEE/CVF international conference on computer vision},
  pages={4852--4861},
  year={2019}
}

@article{2025-tpami-qlimit,
 author={Gong, Ruihao and Liu, Xianglong and Li, Yuhang and Fan, Yunqiang and Wei, Xiuying and Guo, Jinyang},
 journal={IEEE Transactions on Pattern Analysis and Machine Intelligence},
 title={Pushing the Limit of Post-Training Quantization},
 year={2025},
 pages={1-15}
}

@article{2025-nn-low-bit-survey,
 title = {A Survey of Low-bit Large Language Models: Basics, Systems, and Algorithms},
 journal = {Neural Networks},
 pages = {107856},
 year = {2025},
 issn = {0893-6080},
 author = {Gong, Ruihao and Ding, Yifu and Wang, Zining and Lv, Chengtao and Zheng, Xingyu and Du, Jinyang and Yong, Yang and Gu, Shiqiao and Qin, Haotong and Guo, Jinyang and Lin, Dahua and Magno, Michele and Liu, Xianglong},
}

@inproceedings{2021-iclr-brecq,
 title={BRECQ: Pushing the Limit of Post-Training Quantization by Block Reconstruction},
 author={Yuhang Li and Ruihao Gong and Xu Tan and Yang Yang and Peng Hu and Qi Zhang and Fengwei Yu and Wei Wang and Shi Gu},
 booktitle={International Conference on Learning Representations},
 year={2021}
}
\bibliographystyle{icml2026}

%%%%%%%%%%%%%%%%%%%%%%%%%%%%%%%%%%%%%%%%%%%%%%%%%%%%%%%%%%%%%%%%%%%%%%%%%%%%%%%
%%%%%%%%%%%%%%%%%%%%%%%%%%%%%%%%%%%%%%%%%%%%%%%%%%%%%%%%%%%%%%%%%%%%%%%%%%%%%%%
% APPENDIX
%%%%%%%%%%%%%%%%%%%%%%%%%%%%%%%%%%%%%%%%%%%%%%%%%%%%%%%%%%%%%%%%%%%%%%%%%%%%%%%
%%%%%%%%%%%%%%%%%%%%%%%%%%%%%%%%%%%%%%%%%%%%%%%%%%%%%%%%%%%%%%%%%%%%%%%%%%%%%%%
\newpage
\appendix
\onecolumn
\section{Implementation details of Baselines}

Since most existing sparse attention methods are originally designed for bidirectional video generation models, applying them to autoregressive video generation requires additional clarification and careful consideration.

\begin{itemize}
    \item STA~\cite{zhang2025fast}: STA partitions tokens into 3D tiles and applies sparse attention to neighboring tiles. In all experiments, we use a window size of $(3,3,3)$. The original paper keeps early timesteps in dense attention. Since autoregressive models are typically few-step generators (\eg, 4 steps), we do not adopt this setting and apply sparse attention to all steps.
    \item Radial Attention~\cite{li2025radial}: Since the key-value (KV) sequence length varies over chunks in autoregressive video generation, the effective sparsity ratio of Radial Attention also changes accordingly. For 5\,s videos, we perform inference over 7 chunks (3 frames per chunk) with the following sparsity ratios: 67.7, 76.9, 80.6, 82.4, 83.5, 84.9, 86.5.
    \item SVG2~\cite{yang2025sparse}: Since SVG2 relies on K-means clustering and the sequence length in autoregressive generators is shorter than that in bidirectional models, we adjust several hyperparameters to improve its runtime efficiency: \texttt{num\_q\_centroids}=50, \texttt{num\_k\_centroids}=100, \texttt{kmeans\_iter\_init}=20, \texttt{top\_p\_kmeans}=0.9, \texttt{min\_kc\_ratio}=0.10, and \texttt{kmeans\_iter\_step}=2. Notably, whenever the KV length changes in AR models, we re-initialize K-means clustering accordingly.
    \item \textsc{Light Forcing}: 1) We provide detailed duration-specific settings for both short- and long-video generation in Tab.~\ref{tab:lf_hyperparams}, including the target sparsity ratio, $n_{\text{past\_keep}}$ (the number of retained past frames), $n_{\text{sink}}$ (the number of retained earliest historical frames), $n_{\text{win}}$ (the number of retained nearest historical frames), and the corresponding open-source model links. 2) HSA is activated only when the number of historical frames is larger than $n_{\text{past\_keep}}$. Otherwise, all historical frames are preserved. 3) As discussed in the main paper, we keep dense attention for the first chunk and apply Chunk-Aware Growth for sparsity allocation in subsequent chunks. This is because the first chunk has a relatively short sequence length, so sparse attention yields limited speedup but can cause pronounced performance degradation.

\end{itemize}

\begin{table}[ht]
  \setlength{\tabcolsep}{1.5em}
  \centering
  \caption{Hyperparameter settings of \textsc{Light Forcing} for 5-second and 15-second video generation.}
  \resizebox{0.6\linewidth}{!}{
  \begin{tabular}{lcccccc}
    \toprule
    Duration & $s_{\text{target}}$ & $s_{\text{base}}$ & $n_{\text{past\_keep}}$ & $n_{\text{sink}}$ & $n_{\text{win}}$ & Model \\
    \midrule
    5s  & 0.88 & 0.98 & 6 & 1 & 2 & \href{https://huggingface.co/mack-williams/Light-Forcing/blob/main/short_video_gen.pt}{link} \\
    15s & 0.85 & 0.95 & 3 & 1 & 1 & \href{https://huggingface.co/mack-williams/Light-Forcing/blob/main/long_video_gen.pt}{link} \\
    \bottomrule
  \end{tabular}}
  \label{tab:lf_hyperparams}
\end{table}

\section{Prompts for Long Video Generation}
The prompts used for the qualitative examples of 15-second long video generation in the main paper are listed below.

\noindent\textbf{Prompt 1.}
A 3D animation of a small, round, fluffy creature with big, expressive eyes exploring a vibrant, enchanted forest. The creature, a whimsical blend of a rabbit and a squirrel, has soft blue fur and a bushy, striped tail. It hops along a sparkling stream, its eyes wide with wonder. The forest is alive with magical elements: flowers that glow and change colors, trees with leaves in shades of purple and silver, and small floating lights that resemble fireflies. The creature stops to interact playfully with a group of tiny, fairy-like beings dancing around a mushroom ring. The creature looks up in awe at a large, glowing tree that seems to be the heart of the forest. The scene is rendered in a detailed, fantasy style, with a soft, ethereal lighting that enhances the enchantment. The camera follows the creature as it moves, capturing its playful interactions and the magical ambiance of the forest. A medium shot with a dynamic angle that highlights the creature's expressions and the enchanting environment.

\noindent\textbf{Prompt 2.}
An astronaut in a sleek, white spacesuit walks between two ancient stone buildings, their surfaces adorned with intricate carvings and moss. The astronaut's helmet reflects the dim, otherworldly light casting shadows across the worn stones. The buildings loom large, creating a narrow passage that seems to stretch into the distance. The background shows a barren landscape with distant, rocky hills and a pale, orange sky. The astronaut moves with a determined gait, one hand on the building's surface, the other holding a small device. The photo has a realistic, high-resolution texture, capturing the astronaut's focused expression and the textures of the ancient architecture. A medium shot from a slightly elevated angle, emphasizing the contrast between the modern astronaut and the ancient structures.

\section{Theoretical proof of CAG}
\textbf{Denoising-with-re-noising Markov kernel (chunk-wise).}
Fix an AR chunk index $i$ and conditioning $(\boldsymbol{x}^{<i},c)$, and let the inference schedule be
$t_T>t_{T-1}>\cdots>t_0$ with corresponding noise levels $\{\sigma_{t_j}\}_{j=0}^T\subset(0,1]$.
The transition operator $\Psi$ induces the stochastic update
\begin{equation}
\boldsymbol{x}^{i}_{t_{j-1}}
=
\Psi\!\Big(
G_\theta(\boldsymbol{x}^{i}_{t_j},\, t_j,\, \boldsymbol{x}^{<i},\, c),
\, \boldsymbol{\epsilon}_{t_{j-1}},\, t_{j-1}
\Big)
=
(1-\sigma_{t_{j-1}})\,G_\theta(\boldsymbol{x}^{i}_{t_j}, t_j, \boldsymbol{x}^{<i}, c)
+\sigma_{t_{j-1}}\,\boldsymbol{\epsilon}_{t_{j-1}},
\label{eq:ar-renoise-kernel}
\end{equation}
where $\boldsymbol{\epsilon}_{t_{j-1}}\sim\mathcal N(\mathbf 0,\mathbf I)$ are i.i.d. across steps.
Hence, conditional on $\boldsymbol{x}^{i}_{t_j}$, the transition is Gaussian:
\begin{equation}
\boldsymbol{x}^{i}_{t_{j-1}} \mid \boldsymbol{x}^{i}_{t_j}
\sim
\mathcal N\!\big(\boldsymbol{\mu}_{\theta,j}(\boldsymbol{x}^{i}_{t_j}),\; \sigma_{t_{j-1}}^2\mathbf I\big),
\qquad
\boldsymbol{\mu}_{\theta,j}(y):=(1-\sigma_{t_{j-1}})\,G_\theta(y,t_j,\boldsymbol{x}^{<i},c).
\label{eq:gaussian-kernel}
\end{equation}

\textbf{Ideal reverse kernel and mean-map error.}
Let $\boldsymbol{\mu}^\star_j(\cdot)$ be the \emph{ideal} reverse mean map (defined by the exact score / optimal denoiser
under the same schedule), and denote by $p_0(\cdot\mid \boldsymbol{x}^{<i},c)$ the true conditional data distribution
of $\boldsymbol{x}^i$.
Let $q_0(\cdot\mid \boldsymbol{x}^{<i},c)$ be the distribution of the generated output after $T$ transitions from
$\boldsymbol{x}^i_{t_T}\sim\mathcal N(\mathbf 0,\mathbf I)$.
Assume the (average) conditional mean-map error
\begin{equation}
\frac{1}{T}\sum_{j=1}^T
\mathbb E\Big[
\big\|
\boldsymbol{\mu}_{\theta,j}(\boldsymbol{x}^{i}_{t_j})
-\boldsymbol{\mu}^\star_j(\boldsymbol{x}^{i}_{t_j})
\big\|_2^2
\;\Big|\; \boldsymbol{x}^{<i},c
\Big]
\;\le\;
\varepsilon_{\mathrm{mean}}^2,
\label{eq:mean-map-error}
\end{equation}
which is implied by the score/denoiser estimation accuracy as in the assumptions of Theorem~3 in
\cite{li2023towards}.

\textbf{Stepwise KL control (Gaussian KL).}
For Gaussians with equal covariance, we have
\begin{equation}
\mathrm{KL}\!\left(
\mathcal N(\boldsymbol{\mu}^\star,\Sigma)\,\|\,\mathcal N(\boldsymbol{\mu},\Sigma)
\right)
=
\frac12\big\|\Sigma^{-1/2}(\boldsymbol{\mu}^\star-\boldsymbol{\mu})\big\|_2^2.
\label{eq:gaussian-kl}
\end{equation}
Using $\Sigma=\sigma_{t_{j-1}}^2\mathbf I$ in \eqref{eq:gaussian-kernel} yields the per-step contribution
\begin{equation}
\mathbb E\!\left[
\mathrm{KL}\!\left(
\mathcal N(\boldsymbol{\mu}^\star_j(\boldsymbol{x}^{i}_{t_j}),\sigma_{t_{j-1}}^2\mathbf I)
\;\Big\|\;
\mathcal N(\boldsymbol{\mu}_{\theta,j}(\boldsymbol{x}^{i}_{t_j}),\sigma_{t_{j-1}}^2\mathbf I)
\right)
\;\Big|\;\boldsymbol{x}^{<i},c
\right]
=
\frac{1}{2\sigma_{t_{j-1}}^2}\,
\mathbb E\!\left[
\big\|
\boldsymbol{\mu}_{\theta,j}(\boldsymbol{x}^{i}_{t_j})
-\boldsymbol{\mu}^\star_j(\boldsymbol{x}^{i}_{t_j})
\big\|_2^2
\;\Big|\;\boldsymbol{x}^{<i},c
\right].
\label{eq:per-step-kl}
\end{equation}

\textbf{Telescoping KL and TV bound.}
Applying the KL chain rule along the Markov chain induced by \eqref{eq:ar-renoise-kernel}, one obtains
\begin{equation}
\mathrm{KL}\!\left(q_0(\cdot\mid \boldsymbol{x}^{<i},c)\,\|\,p_0(\cdot\mid \boldsymbol{x}^{<i},c)\right)
\;\le\;
\sum_{j=1}^T
\frac{1}{2\sigma_{t_{j-1}}^2}\,
\mathbb E\!\left[
\big\|
\boldsymbol{\mu}_{\theta,j}(\boldsymbol{x}^{i}_{t_j})
-\boldsymbol{\mu}^\star_j(\boldsymbol{x}^{i}_{t_j})
\big\|_2^2
\;\Big|\;\boldsymbol{x}^{<i},c
\right]
\;+\;
\mathrm{KL}\!\left(q_{t_T}\,\|\,p_{t_T}\right).
\label{eq:kl-telescope-ar}
\end{equation}
By Pinsker's inequality,
\begin{equation}
\mathrm{TV}\!\left(q_0(\cdot\mid \boldsymbol{x}^{<i},c),\,p_0(\cdot\mid \boldsymbol{x}^{<i},c)\right)
\le
\sqrt{\frac12\,
\mathrm{KL}\!\left(q_0(\cdot\mid \boldsymbol{x}^{<i},c)\,\|\,p_0(\cdot\mid \boldsymbol{x}^{<i},c)\right)}.
\label{eq:pinsker-ar}
\end{equation}

\textbf{Error vs.\ number of denoising steps $T$.}
Under the same regularity and schedule conditions in \cite{li2023towards},
the discretization/mixing term contributes $\tilde{\mathcal O}(d/T)$ in KL (polylog factors),
while the model (score) error contributes $\tilde{\mathcal O}(d\,\varepsilon_{\mathrm{score}}^2)$ in KL.
Combining \eqref{eq:pinsker-ar} with Eq.~(31) of \cite{li2023towards} gives the conditional TV guarantee
\begin{equation}
\mathrm{TV}\!\left(q_0(\cdot\mid \boldsymbol{x}^{<i},c),\,p_0(\cdot\mid \boldsymbol{x}^{<i},c)\right)
\;\le\;
C_1\cdot \frac{d\,\log^3 T}{\sqrt{T}}
\;+\;
C_2\cdot \sqrt{d}\,\varepsilon_{\mathrm{score}}\,\log^2 T,
\label{eq:tv-vs-T-ar}
\end{equation}
for universal constants $C_1,C_2>0$. In particular, when $\varepsilon_{\mathrm{score}}=0$, the sampling error decays as
$\tilde{\mathcal O}(d/\sqrt{T})$, whereas for imperfect models it saturates at
$\tilde{\mathcal O}(\sqrt{d}\,\varepsilon_{\mathrm{score}})$ as $T\to\infty$.

\section{Detailed VBench Results}
We report the full VBench~\cite{huang2024vbench} results across all dimensions for each method in Tab.~\ref{tab:quality} and Tab.~\ref{tab:semantic}.

\begin{table*}[ht!]\setlength{\tabcolsep}{1pt}
 \renewcommand{\arraystretch}{1.05}
  \centering

  % \caption{Performance comparison with relevant baselines on $8$ dimensions of VBench~\cite{huang2024vbench}. ``+DMD2'' denotes our 4-step distilled \textsc{LinVideo} model. We highlight the best score and the second score in \textbf{bold} and \underline{underlined} formats, respectively.} 
   \caption{Detailed performance comparison with state-of-the-art baselines on VBench~\cite{huang2024vbench} (quality part).} 
   % . Full results are provided in the Appendix.
  \resizebox{0.8\linewidth}{!}{
  \begin{tabular}[t!]{l|ccccccc}
\toprule
Method 
& \rotatebox{0}{\makecell{Subject\\Consistency}$\uparrow$}
& \rotatebox{0}{\makecell{Background\\Consistency}$\uparrow$}
& \rotatebox{0}{\makecell{Temporal\\Flickering}$\uparrow$}
& \rotatebox{0}{\makecell{Motion\\Smoothness}$\uparrow$}
& \rotatebox{0}{\makecell{Aesthetic\\Quality}$\uparrow$}
& \rotatebox{0}{\makecell{Imaging\\Quality}$\uparrow$}
& \rotatebox{0}{\makecell{Dynamic\\Degree}$\uparrow$} \\
\midrule
\multicolumn{8}{c}{\cellcolor[gray]{0.92}Self-Forcing 1.3B ($\texttt{fps}=16$)} \\
\midrule
FlashAttention2~\cite{dao2023flashattention} & 95.3 & 96.5 & 99.1 & 98.3 & 67.4 & 70.0 & 63.1 \\
STA~\cite{zhang2025fast}              & 96.3 & 96.9 & 99.2 & 98.5 & 64.5 & 71.7 & 48.9 \\
Radial~\cite{li2025radial}           & 90.2 & 93.6 & 95.6 & 96.0 & 45.8 & 66.1 & 88.6 \\
SVG2~\cite{yang2025sparse}             & 93.6 & 95.6 & 98.2 & 97.8 & 66.0 & 68.2 & 72.8 \\
SLA~\cite{zhang2025sla}              & 95.6 & 96.7 & 99.2 & 98.3 & 66.7 & 69.8 & 44.2 \\
VMoBA~\cite{wu2025vmoba}            & 92.8 & 95.5 & 98.0 & 97.3 & 65.2 & 69.9 & 84.2 \\
\rowcolor{mycolor!30}\textsc{Light Forcing}
                 & \textbf{96.2} & 96.5 & \textbf{99.2} & 98.3 & \textbf{67.2} & \textbf{71.0} & 66.7 \\
\midrule
\multicolumn{8}{c}{\cellcolor[gray]{0.92}LongLive 1.3B ($\texttt{fps}=16$)} \\
\midrule
FlashAttention2~\cite{dao2023flashattention} & 97.0 & 97.2 & 99.3 & 98.8 & 68.7 & 69.3 & 39.2 \\
STA~\cite{zhang2025fast}              & 97.4 & 97.8 & 99.6 & 99.0 & 65.6 & 71.2 & 22.8 \\
Radial~\cite{li2025radial}           & 77.6 & 88.9 & 98.1 & 98.0 & 55.1 & 72.0 & 25.0 \\
SVG2~\cite{yang2025sparse}              & 95.3 & 96.1 & 98.8 & 98.5 & 66.7 & 67.0 & 44.4 \\
VMoBA~\cite{wu2025vmoba}            & 58.3 & 80.9 & 97.6 & 97.5 & 59.9 & 68.2 & 50.6 \\
\rowcolor{mycolor!30}\textsc{Light Forcing}
                 & \textbf{96.9} & 96.7 & 98.9 & 98.2 & \textbf{67.2} & \textbf{70.6} & \textbf{59.4} \\
\bottomrule
\end{tabular}
}
    \label{tab:quality}
\end{table*}

\begin{table*}[ht!]\setlength{\tabcolsep}{1pt}
 \renewcommand{\arraystretch}{1.05}
  \centering

  % \caption{Performance comparison with relevant baselines on $8$ dimensions of VBench~\cite{huang2024vbench}. ``+DMD2'' denotes our 4-step distilled \textsc{LinVideo} model. We highlight the best score and the second score in \textbf{bold} and \underline{underlined} formats, respectively.} 
   \caption{Detailed performance comparison with state-of-the-art baselines on VBench~\cite{huang2024vbench} (semantic part).} 
   % . Full results are provided in the Appendix.
  \resizebox{0.8\linewidth}{!}{
  \begin{tabular}[t!]{l|ccccccccc}
\toprule
Method 
& \rotatebox{0}{\makecell{Object\\Class}$\uparrow$}
& \rotatebox{0}{\makecell{Multiple\\Objects}$\uparrow$}
& \rotatebox{0}{\makecell{Human\\Action}$\uparrow$}
& \rotatebox{0}{\makecell{Color}$\uparrow$}
& \rotatebox{0}{\makecell{Spatial\\Relationship}$\uparrow$}
& \rotatebox{0}{\makecell{Scene}$\uparrow$}
& \rotatebox{0}{\makecell{Appearance\\Style}$\uparrow$}
& \rotatebox{0}{\makecell{Temporal\\Style}$\uparrow$}
& \rotatebox{0}{\makecell{Overall\\Consistency}$\uparrow$} \\
\midrule
\multicolumn{10}{c}{\cellcolor[gray]{0.92}Self-Forcing 1.3B ($\texttt{fps}=16$)} \\
\midrule
FlashAttention2~\cite{dao2023flashattention} & 94.9 & 88.4 & 96.4 & 88.6 & 83.1 & 54.4 & 20.6 & 24.6 & 26.9 \\
STA~\cite{zhang2025fast}              & 95.2 & 86.1 & 95.2 & 91.7 & 91.1 & 57.1 & 22.1 & 23.0 & 25.5 \\
Radial~\cite{li2025radial}           & 56.0 & 31.8 & 80.4 & 86.5 & 39.0 & 15.1 & 22.3 & 15.9 & 18.1 \\
SVG2~\cite{yang2025sparse}             & 93.5 & 73.5 & 96.4 & 87.8 & 76.9 & 54.2 & 20.4 & 24.5 & 27.0 \\
SLA~\cite{zhang2025sla}              & 96.4 & 87.5 & 96.8 & 91.8 & 89.3 & 56.3 & 20.5 & 24.2 & 26.8 \\
VMoBA~\cite{wu2025vmoba}            & 93.9 & 81.1 & 96.8 & 90.5 & 81.0 & 55.3 & 20.5 & 24.3 & 26.8 \\
\rowcolor{mycolor!30}\textsc{Light Forcing}
                 & 94.3 & \textbf{88.9} & 96.0 & 88.2 & 81.4 & 55.3 & 20.1 & 24.6 & \textbf{26.9} \\
\midrule
\multicolumn{10}{c}{\cellcolor[gray]{0.92}LongLive 1.3B ($\texttt{fps}=16$)} \\
\midrule
FlashAttention2~\cite{dao2023flashattention} & 94.4 & 87.8 & 96.4 & 89.2 & 78.5 & 55.6 & 20.6 & 24.3 & 26.7 \\
STA~\cite{zhang2025fast}              & 95.5 & 86.0 & 95.4 & 94.7 & 90.4 & 54.8 & 21.7 & 22.2 & 25.1 \\
Radial~\cite{li2025radial}           & 72.7 & 59.9 & 93.2 & 93.0 & 66.6 & 16.7 & 22.4 & 19.4 & 22.4 \\
SVG2~\cite{yang2025sparse}             & 92.3 & 78.9 & 96.0 & 88.5 & 74.5 & 53.9 & 20.4 & 24.4 & 27.0 \\
VMoBA~\cite{wu2025vmoba}            & 78.2 & 64.8 & 96.4 & 89.3 & 64.6 & 26.8 & 21.5 & 23.2 & 26.0 \\
\rowcolor{mycolor!30}\textsc{Light Forcing}
                 & \textbf{95.3} & \textbf{89.6} & 96.6 & 86.1 & 78.7 & 53.3 & 20.1 & 24.4 & \textbf{26.7} \\
\bottomrule
\end{tabular}
}
    \label{tab:semantic}
\end{table*}

\begin{table*}[t!]
\renewcommand{\arraystretch}{1.08}
\centering
\caption{Comparison with feature-cache and token-reduction acceleration methods on latency and representative VBench~\cite{huang2024vbench} metrics.}
% \vspace{2pt}
\footnotesize
\resizebox{0.92\linewidth}{!}{
\begin{tabular}[t]{l|cc|cccccc}
\toprule
\multirow{2}{*}{Method} & \multirow{2}{*}{Latency~(s)$\downarrow$} & \multirow{2}{*}{Speedup$\uparrow$} & \multirow{2}{*}{\rotatebox{0}{\makecell{Subject\\Consistency}$\uparrow$}} & \multirow{2}{*}{\rotatebox{0}{\makecell{Background\\Consistency}$\uparrow$}} & \multirow{2}{*}{\rotatebox{0}{\makecell{Aesthetic\\Quality}$\uparrow$}} & \multirow{2}{*}{\rotatebox{0}{\makecell{Imaging\\Quality}$\uparrow$}} & \multirow{2}{*}{\rotatebox{0}{\makecell{Multiple\\Objects}$\uparrow$}} & \multirow{2}{*}{\rotatebox{0}{\makecell{Scene}$\uparrow$}} \\
& & & & & & & & \\
\midrule
TeaCache~\cite{liu2025timestep} & 8.32 & $1.15\times$ & 94.9 & 96.1 & 67.0 & 69.5 & 84.4 & 52.2 \\
DyCoke~\cite{tao2025dycoke} & 8.71 & $1.10\times$ & 94.5 & 95.6 & 67.2 & 69.5 & 87.5 & 54.2 \\
\midrule
\rowcolor{mycolor!30}\textsc{Light Forcing} & \textbf{7.39} & $\mathbf{1.30}\times$ & \textbf{96.2} & \textbf{96.5} & \textbf{67.2} & \textbf{71.0} & \textbf{88.9} & \textbf{55.3} \\
\bottomrule
\end{tabular}
}
\label{tab:other_method}
\end{table*}

\section{Detailed VBench-Long Results}
We report the full VBench-Long~\cite{huang2025vbench++} results for 15-second generation in Tab.~\ref{tab:long_detail}.

\begin{table*}[ht!]\setlength{\tabcolsep}{1pt}
 \renewcommand{\arraystretch}{1.05}
  \centering
   \caption{Detailed long-video generation comparison on VBench-Long~\cite{huang2025vbench++}.}
  \resizebox{0.9\linewidth}{!}{
  \begin{tabular}[t!]{l|cccccccc}
\toprule
Method 
& \rotatebox{0}{\makecell{Subject\\Consistency}$\uparrow$}
& \rotatebox{0}{\makecell{Background\\Consistency}$\uparrow$}
& \rotatebox{0}{\makecell{Temporal\\Flickering}$\uparrow$}
& \rotatebox{0}{\makecell{Motion\\Smoothness}$\uparrow$}
& \rotatebox{0}{\makecell{Aesthetic\\Quality}$\uparrow$}
& \rotatebox{0}{\makecell{Imaging\\Quality}$\uparrow$}
& \rotatebox{0}{\makecell{Dynamic\\Degree}$\uparrow$}
& \rotatebox{0}{\makecell{Object\\Class}$\uparrow$} \\
\midrule
\multicolumn{9}{c}{\cellcolor[gray]{0.92}Infinite-Forcing 1.3B ($\texttt{fps}=16$)} \\
\midrule
FlashAttention2~\cite{dao2023flashattention} & 98.3 & 97.6 & 99.5 & 98.5 & 65.0 & 68.7 & 54.7 & 95.1 \\
\rowcolor{mycolor!30}\textsc{Light Forcing}
                 & 98.0 & 97.0 & 99.4 & 98.6 & 65.1 & 69.5 & 64.7 & 95.4 \\
\midrule
Method 
& \rotatebox{0}{\makecell{Multiple\\Objects}$\uparrow$}
& \rotatebox{0}{\makecell{Human\\Action}$\uparrow$}
& \rotatebox{0}{\makecell{Color}$\uparrow$}
& \rotatebox{0}{\makecell{Spatial\\Relationship}$\uparrow$}
& \rotatebox{0}{\makecell{Scene}$\uparrow$}
& \rotatebox{0}{\makecell{Appearance\\Style}$\uparrow$}
& \rotatebox{0}{\makecell{Temporal\\Style}$\uparrow$}
& \rotatebox{0}{\makecell{Overall\\Consistency}$\uparrow$} \\
\midrule
\multicolumn{9}{c}{\cellcolor[gray]{0.92}Infinite-Forcing 1.3B ($\texttt{fps}=16$)} \\
\midrule
FlashAttention2~\cite{dao2023flashattention} & 84.7 & 96.1 & 86.2 & 74.0 & 57.8 & 20.7 & 24.1 & 26.1 \\
\rowcolor{mycolor!30}\textsc{Light Forcing}
                 & 85.1 & 94.5 & 85.3 & 77.5 & 52.7 & 20.5 & 24.6 & 26.2 \\
\bottomrule
\end{tabular}
}
    \label{tab:long_detail}
\end{table*}

\section{Qualitative Ablation Study}
\begin{figure*}[ht!]
   \centering
        \includegraphics[width=0.9\textwidth]{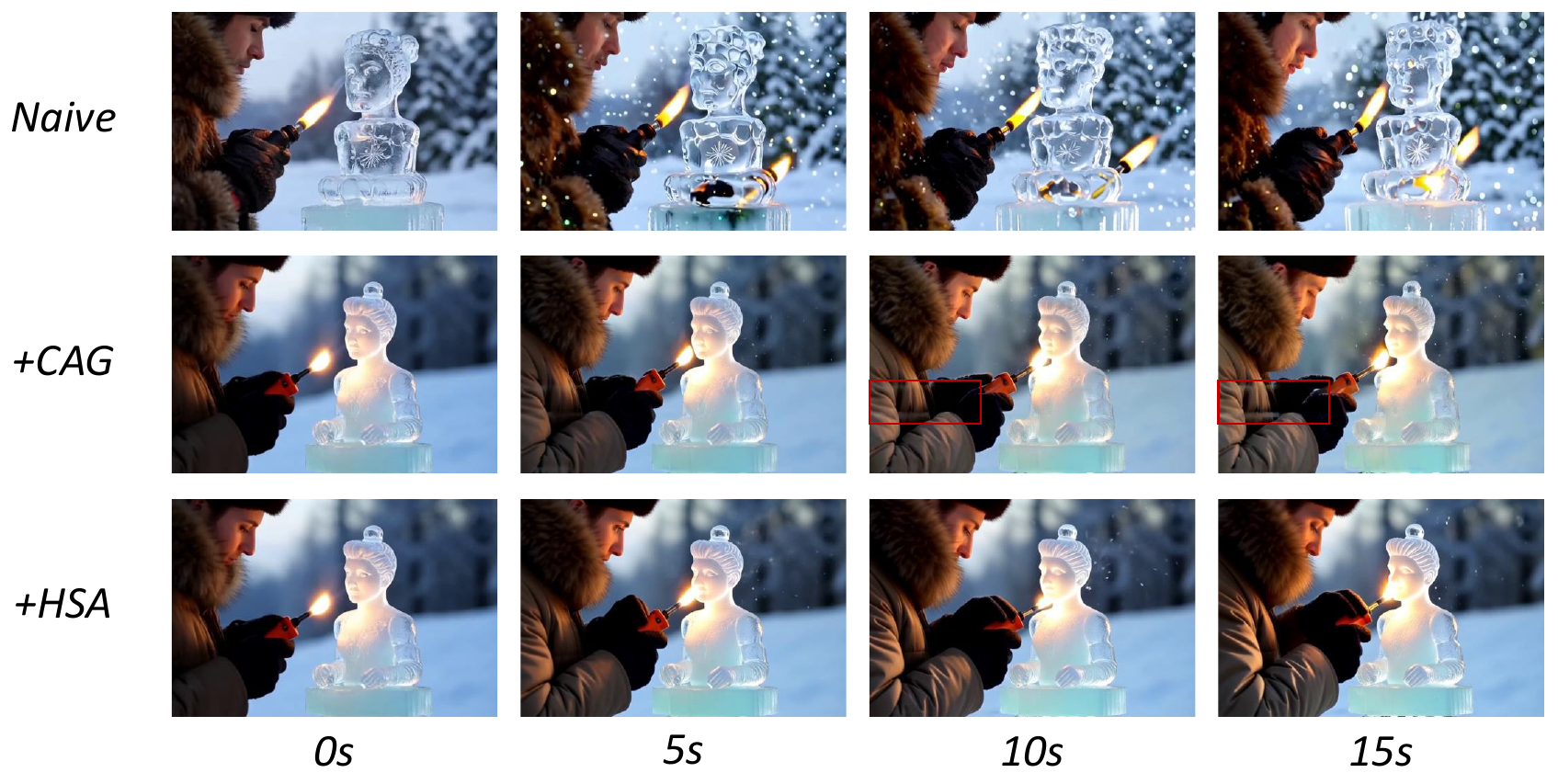}
        \caption{Qualitative ablation study of CAG and HSA for long-video generation on the first example.}
        \label{fig:ablation_supp1}
\end{figure*}

\begin{figure*}[ht!]
   \centering
        \includegraphics[width=0.9\textwidth]{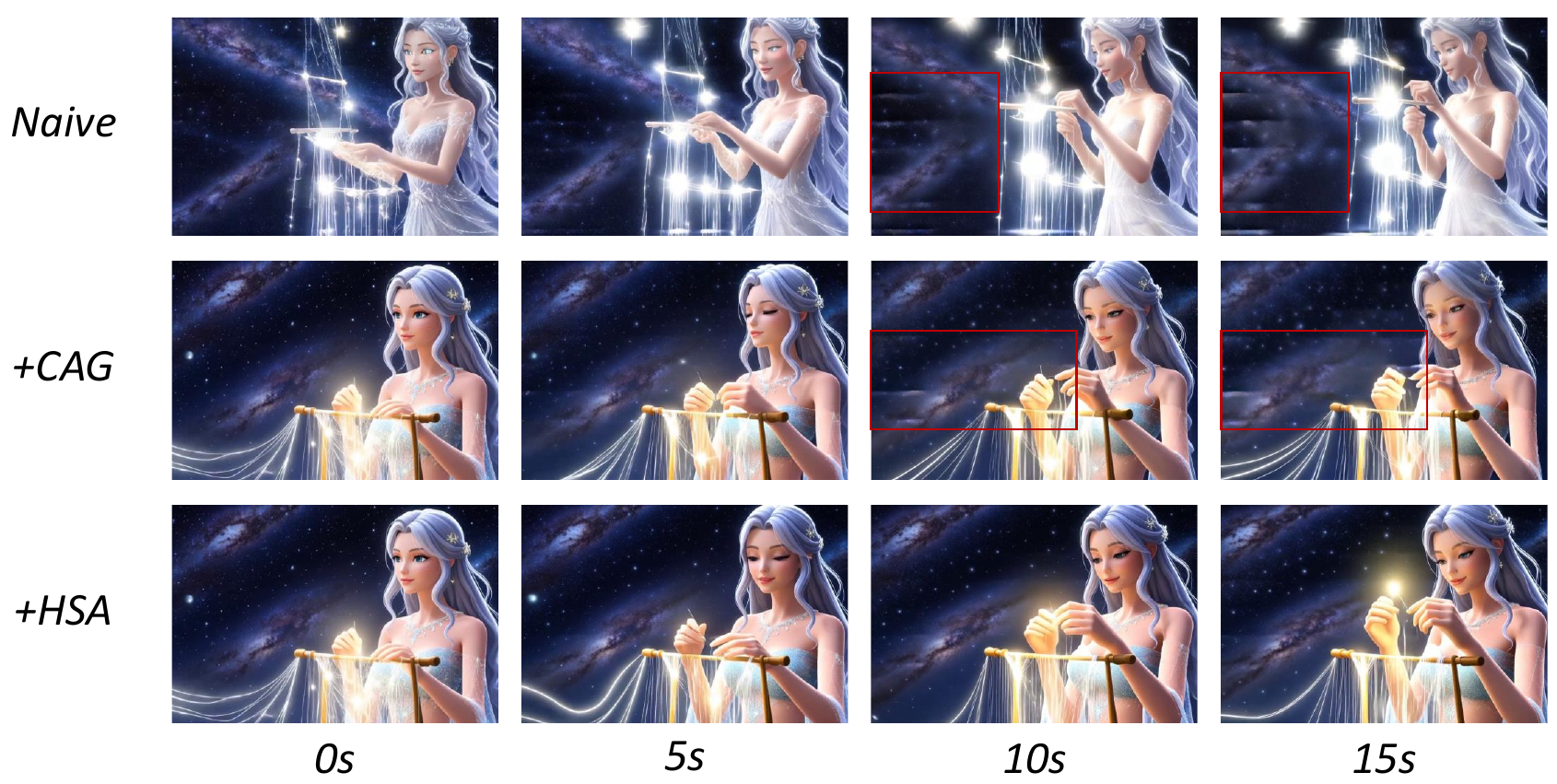}
        \caption{Qualitative ablation study of CAG and HSA for long-video generation on the second example.}
        \label{fig:ablation_supp2}
\end{figure*}

We further provide long-video qualitative ablations for the two proposed modules in \textsc{Light Forcing}, \ie, Chunk-Aware Growth (CAG) and Hierarchical Sparse Attention (HSA). All three variants use the trained long-video version of \textsc{Light Forcing}\footnote{\url{https://huggingface.co/mack-williams/Light-Forcing/blob/main/long_video_gen.pt}}. For the naive baseline, we adopt a fixed sparsity ratio of 85\% across chunks. As shown in Fig.~\ref{fig:ablation_supp1} and Fig.~\ref{fig:ablation_supp2}, the naive strategy suffers from severe color degradation, and the first example exhibits noticeable noise starting from around 5s. Adding CAG largely alleviates the color degradation by allocating the sparsity budget according to chunk-wise generation difficulty, but white band-like artifacts still appear in the latter part of the videos (\eg, around 10s). After further incorporating HSA, these artifacts are effectively mitigated, leading to more stable long-video generation with better visual quality and temporal consistency.

\section{Comparison with Other Acceleration Methods}
We further compare \textsc{Light Forcing} with two representative acceleration paradigms beyond sparse attention~\cite{yang2021r3det,yang2019deep,li2018self,deng2018triplet,gong2019differentiable,2025-tpami-qlimit,2025-nn-low-bit-survey,2021-iclr-brecq}: feature caching and token reduction. TeaCache~\cite{liu2025timestep} estimates the change between model outputs from timestep embeddings and uses this signal to decide when intermediate model outputs can be reused during denoising. DyCoke~\cite{tao2025dycoke} combines temporal token merging with dynamic KV-cache pruning, aiming to remove redundant tokens while preserving visually important information.

For implementation, we set the TeaCache threshold to $0.2$. For DyCoke, we follow the LLMC+~\cite{lv2026llmc+} implementation and adapt it to the chunk-wise generation protocol of Self Forcing~\cite{huang2025self}. Specifically, we group three consecutive frames as one unit, matching the three-frame chunk used by Self Forcing. Within each group, we keep all tokens from the first frame and retain only $25\%$ of the tokens from the second and third frames, resulting in a total historical KV-cache retention ratio of $50\%$.

As shown in Tab.~\ref{tab:other_method}, under the highest-speedup setting among these methods ($1.30\times$), \textsc{Light Forcing} achieves the best accuracy on all reported representative metrics. It obtains higher subject consistency ($96.2$ vs. $94.9/94.5$), background consistency ($96.5$ vs. $96.1/95.6$), imaging quality ($71.0$ vs. $69.5/69.5$), multiple-object reasoning ($88.9$ vs. $84.4/87.5$), and scene understanding ($55.3$ vs. $52.2/54.2$) than TeaCache and DyCoke, respectively. These results suggest that directly exploiting the autoregressive attention structure preserves visual fidelity more effectively than reusing intermediate features or aggressively reducing historical tokens. We also observe in qualitative comparisons that TeaCache tends to introduce blurry frames, while DyCoke often causes abrupt motion changes due to temporal token removal.

\section{More Efficiency Analysis}

\subsection{More Devices}
To further assess the hardware generality of \textsc{Light Forcing}, we evaluate the same inference pipeline on three representative GPU platforms: RTX~5090, A100, and H100. All measurements use a single GPU, record the generation time of one video after operator warm-up, and start timing from the second generated sample. Unless otherwise specified, all experiments are conducted with the same Docker image\footnote{\texttt{lvchengtao/light\_forcing:v1}.}, which provides the optimized sparse-attention, FP8, RoPE, RMSNorm, and related deployment kernels used in our implementation. On RTX~5090 and A100, \textsc{Light Forcing} invokes the Triton sparse-attention kernel, while on H100 it uses the FlashAttention 4~\cite{zadouri2026flashattention} sparse kernel.

\begin{table*}[ht!]\setlength{\tabcolsep}{5pt}
\renewcommand{\arraystretch}{1.16}
\centering
\caption{Additional efficiency results across different devices. ``Dense'' denotes the dense-attention baseline on the corresponding device. Each cell reports latency and the end-to-end speedup over the dense baseline under the same duration and device setting.}
\vspace{2pt}
\footnotesize
\resizebox{\linewidth}{!}{
\begin{tabular}[t]{llcccccc}
\toprule
Device & Duration & Dense & +\textsc{Light Forcing} & +FP8 Linear & +Efficient Kernel & +LightVAE & Final Speedup \\
\midrule
RTX~5090 & 5s  & 9.09s (1.00$\times$)  & 6.83s (1.33$\times$) & 5.90s (1.54$\times$) & 5.37s (1.69$\times$) & 2.96s (3.07$\times$) & 3.07$\times$ \\
RTX~5090 & 15s & 30.40s (1.00$\times$) & 24.20s (1.26$\times$) & 21.40s (1.42$\times$) & 17.00s (1.79$\times$) & 9.60s (3.17$\times$) & 3.17$\times$ \\
\midrule
A100 & 5s  & 11.38s (1.00$\times$) & 9.88s (1.15$\times$) & -- & 9.41s (1.21$\times$) & 4.85s (2.35$\times$) & 2.35$\times$ \\
A100 & 15s & 38.28s (1.00$\times$) & 34.56s (1.11$\times$) & -- & 26.63s (1.44$\times$) & 18.08s (2.12$\times$) & 2.12$\times$ \\
\midrule
H100 & 5s  & 4.80s (1.00$\times$)  & 4.33s (1.11$\times$) & 4.32s (1.11$\times$) & 3.74s (1.28$\times$) & 2.39s (2.01$\times$) & 2.01$\times$ \\
H100 & 15s & 15.80s (1.00$\times$) & 14.10s (1.12$\times$) & 13.80s (1.14$\times$) & 12.10s (1.31$\times$) & 8.00s (1.98$\times$) & 1.98$\times$ \\
\bottomrule
\end{tabular}}
\label{tab:more_devices}
\end{table*}

As shown in Tab.~\ref{tab:more_devices}, \textsc{Light Forcing} consistently improves end-to-end latency across consumer and datacenter GPUs. On RTX~5090, sparse attention alone yields $1.33\times$ and $1.26\times$ acceleration for 5s and 15s generation, respectively, and the complete deployment stack reaches $3.07\times$--$3.17\times$ speedup. On H100, although attention computation for the 1.3B model is already relatively lightweight on H-series GPUs, \textsc{Light Forcing} still provides $1.11\times$--$1.12\times$ speedup by itself and nearly $2\times$ acceleration when combined with the remaining deployment optimizations. On A100, where FP8 linear layers are not supported and the sparse kernel has not been specifically tuned for this architecture, \textsc{Light Forcing} still achieves positive gains and reaches $2.12\times$--$2.35\times$ speedup with efficient kernel fusion and LightVAE. These results indicate that \textsc{Light Forcing} is not tied to a single accelerator or kernel implementation, but provides broadly applicable sparse-attention acceleration for autoregressive video diffusion.

\subsection{Peak Memory Analysis}
We also report peak GPU memory in Tab.~\ref{tab:peak_memory_more}. The available peak-memory measurements are collected on RTX~5090 under the same Docker image and inference protocol as above.

\begin{table*}[ht!]
\renewcommand{\arraystretch}{1.22}
\centering
\caption{Peak memory usage on RTX~5090. Values are measured in GB.}
% \vspace{2pt}
% \scriptsize
\resizebox{0.7\linewidth}{!}{
\begin{tabular}[t]{l|ccccc}
\toprule
Duration & Dense & +\textsc{Light Forcing} & +FP8 Linear & +Efficient Kernel & +LightVAE \\
\midrule
5s  & 17.8 & 17.8 & 16.6 & 15.8 & 12.7 \\
15s & 17.6 & 17.6 & 16.5 & 16.3 & 13.1 \\
\bottomrule
\end{tabular}
}
\label{tab:peak_memory_more}
\end{table*}

\textsc{Light Forcing} primarily reduces the number of attention tiles evaluated at runtime, so its main benefit is latency reduction rather than peak-memory reduction in the current implementation. Consequently, the measured peak memory remains unchanged after enabling sparse attention alone. FP8 linear layers and efficient kernels reduce the memory footprint by lowering activation or operator overhead, and LightVAE provides the largest memory reduction by replacing the VAE component with a more efficient variant. Overall, the full deployment stack reduces peak memory from 17.8\,GB to 12.7\,GB for 5s generation and from 17.6\,GB to 13.1\,GB for 15s generation, corresponding to 28.7\% and 25.6\% memory reduction, respectively. The memory profile therefore complements the latency results: \textsc{Light Forcing} supplies broadly applicable attention acceleration, while the accompanying deployment modules further improve the feasibility of running autoregressive video diffusion on memory-constrained devices.

\section{More Visualization Examples}
We provide more detailed qualitative comparisons in the supplementary material. In Fig.~\ref{fig:visual_supp}, we visualize results for all baselines under two prompts, \emph{``A person is clay pottery making''} and \emph{``Turtle swimming in ocean''}. Most baselines exhibit noticeable degradation, including (1) loss of fine-grained details (\eg, distorted hands) and (2) anomalous generations (\eg, a turtle with two heads).

\begin{figure*}[ht!]
   \centering
        \includegraphics[width=0.9\textwidth]{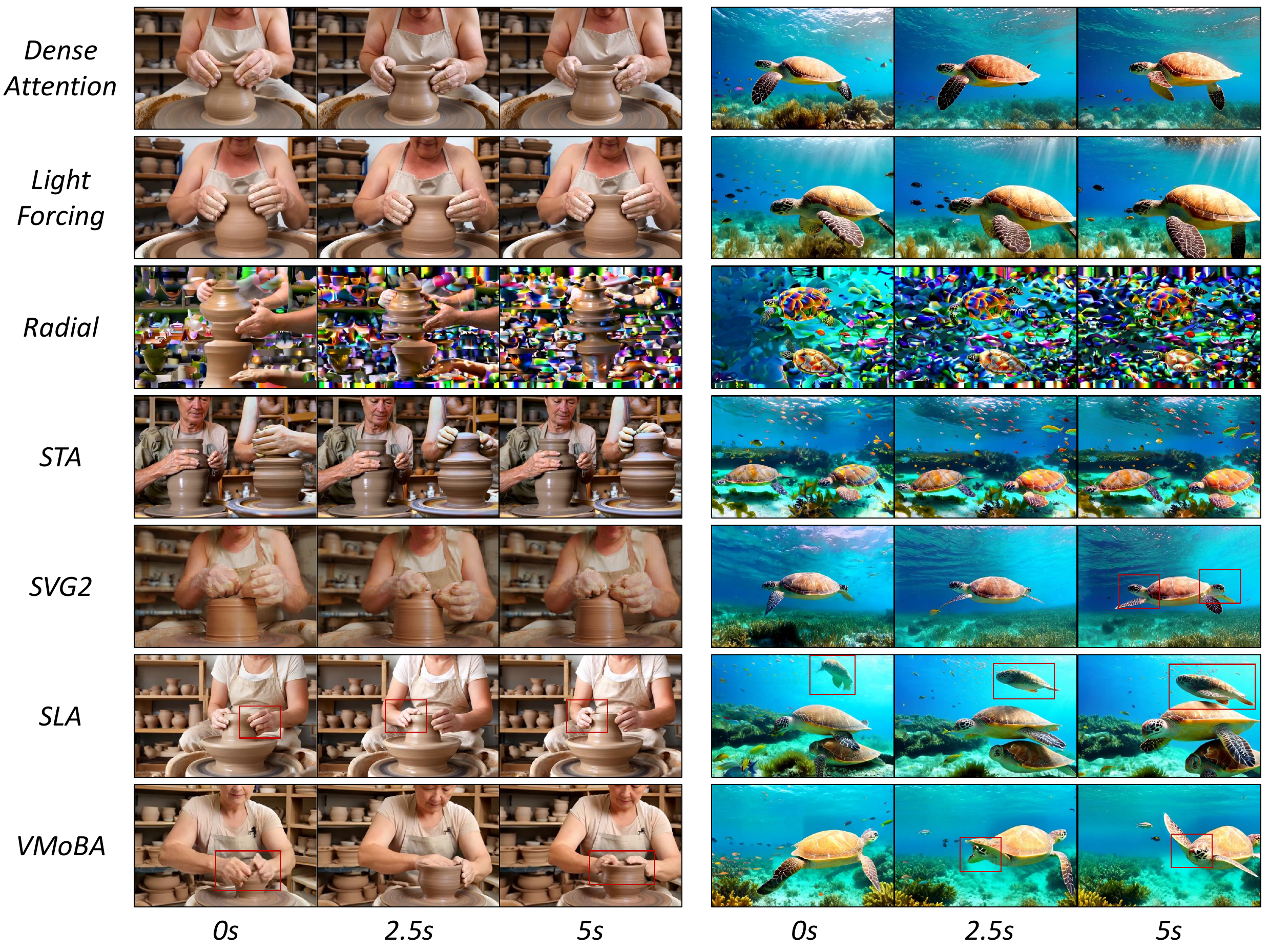}
        \caption{More qualitative examples on Self Forcing~\cite{huang2025self}.}
        \label{fig:visual_supp}
\end{figure*}

\section{Limitations}
Despite the strong empirical results, our work has several limitations. First, we evaluate only 1.3B models. Scaling \textsc{Light Forcing} to larger models (\eg, a 14B realtime-video model~\cite{krea_realtime_14b}) is an important direction. Second, although HSA can alleviate white band-like artifacts, subtle artifacts may still remain in a very small number of samples, which we leave as a future direction. Third, some of our kernel-fusion operators may provide less pronounced acceleration on certain GPUs, such as A100, since they have not been specifically adapted to these architectures. Finally, while sparsity is effective, combining it with other methods (\eg, reducing generation to very few denoising steps, such as 1--3 steps, or low-bit quantization) to push toward extreme acceleration remains open.

% To facilitate a direct qualitative comparison of \textsc{Light Forcing}, we include videos for all baselines in the supplementary material, named in the format \texttt{sample\{id\}\_\{method\_name\}.mp4} (\eg, \texttt{sample1\_svg2.mp4}).

% You can have as much text here as you want. The main body must be at most $8$
% pages long. For the final version, one more page can be added. If you want, you
% can use an appendix like this one.

% The $\mathtt{\backslash onecolumn}$ command above can be kept in place if you
% prefer a one-column appendix, or can be removed if you prefer a two-column
% appendix.  Apart from this possible change, the style (font size, spacing,
% margins, page numbering, etc.) should be kept the same as the main body.
% %%%%%%%%%%%%%%%%%%%%%%%%%%%%%%%%%%%%%%%%%%%%%%%%%%%%%%%%%%%%%%%%%%%%%%%%%%%%%%%
% %%%%%%%%%%%%%%%%%%%%%%%%%%%%%%%%%%%%%%%%%%%%%%%%%%%%%%%%%%%%%%%%%%%%%%%%%%%%%%%

\end{document}